\pdfoutput=1
\documentclass[11pt,a4paper]{article}
\usepackage[T1]{fontenc}
\usepackage[utf8]{inputenc}
\usepackage{lmodern}
\usepackage{microtype}
\usepackage[margin=2.5cm]{geometry}
\usepackage{amsmath,amssymb}
\usepackage{graphicx}
\usepackage{array,tabularx,booktabs}
\usepackage[font=small,labelfont=bf]{caption}
\usepackage{algorithm}
\usepackage{algpseudocode}
\usepackage{fancyvrb}
\usepackage[numbers,sort&compress]{natbib}
\usepackage{authblk}
\usepackage[title,titletoc]{appendix}
\usepackage{xcolor}
\usepackage[colorlinks=true,allcolors=blue!50!black]{hyperref}
\newcolumntype{C}{>{\centering\arraybackslash}X}

\hypersetup{pdftitle={Beyond Accuracy: A Geometric Stability Analysis of Large Language Models in Chess Evaluation},
  pdfauthor={Qingya Hu, Ruifeng Cao, Weiqi Wang, Xidan Song}}

\title{Beyond Accuracy: A Geometric Stability Analysis of Large Language Models in Chess Evaluation}
\author[1,$\dagger$]{Qingya Hu}
\author[2,$\dagger$]{Ruifeng Cao}
\author[3,$\dagger$]{Weiqi Wang}
\author[4,$\dagger$,*]{Xidan Song}
\affil[1]{Department of Computer Science, Wuhan Donghu University, Wuhan 430068, China \protect\\ \texttt{huqingya@wdu.edu.cn}}
\affil[2]{Department of Mechanical and Aerospace Engineering, The University of Manchester, Manchester M13 9PL, UK \protect\\ \texttt{ruifeng.cao@manchester.ac.uk}}
\affil[3]{Department of Computer Science, The University of Manchester, Manchester M13 9PL, UK \protect\\ \texttt{weiqi.wang-2@manchester.ac.uk}}
\affil[4]{School of Computer Science, Wuhan University, Wuhan 430072, China \protect\\ \texttt{liuqingdamafen@whu.edu.cn}}
\affil[$\dagger$]{These authors contributed equally to this work.}
\affil[*]{Corresponding author: \texttt{liuqingdamafen@whu.edu.cn}}
\date{}

\begin{document}
\maketitle

\begin{abstract}
Large language models (LLMs) are usually evaluated in chess by how closely their position evaluations agree with a strong engine, but such agreement does not show that a model evaluates equivalent positions consistently. We apply metamorphic relations to chess evaluation: a $180^\circ$ rotation of pawnless positions, left--right mirroring, color swapping, position pairs one move apart, conversion between Forsyth--Edwards Notation and a natural-language description, and illegal positions. Six LLMs (Claude Sonnet 4.5, DeepSeek Chat, Gemini 2.5 Flash, GPT-5.1, Grok 4.1 Fast and Kimi K2 Turbo) evaluated seven test sets of 500 items each, with Stockfish 16.1 as the reference. All six models failed on rotation: at most 10.6\% of the 500 rotated pairs received evaluations within 300 centipawns (cp) of each other, and the models often declared the legal rotated positions illegal. On the other four transformations, stability ranked the models as accuracy did except for the first two places: GPT-5.1 had the lowest mean stability error (186.17~cp) and Claude Sonnet 4.5 the lowest accuracy error (330.41~cp). Rejection rates for illegal positions ranged from 44.2\% to 96.0\%. Stability should be reported with the share of consistent pairs, because means over answered pairs concealed most rotation failures.

\medskip
\noindent\textbf{Keywords:} large language models; chess evaluation; geometric stability; metamorphic testing; robustness analysis; AI reasoning; evaluation metrics
\end{abstract}
\section{Introduction}\label{sec:intro}

Neural networks are now a core component of game-playing systems. In chess, shogi and Go, AlphaZero reached superhuman performance through self-play~\cite{silver2018general}. High performance in standard tests and competitions does not rule out failures on unusual inputs: adversarial policies trained against the superhuman Go program KataGo defeat it by inducing serious blunders~\cite{wang2023adversarial}. More generally, adversarial examples are inputs with small perturbations that cause a neural network to produce wrong outputs~\cite{goodfellow2015explaining,akhtar2018threat}.

A well-known example of such a failure comes from the 2016 match between AlphaGo~\cite{silver2016mastering} and Lee Sedol. In game~4, Lee's 78th move, which AlphaGo had not anticipated and which commentators called brilliant, was followed by an AlphaGo error and by AlphaGo's only loss in the five-game match, which it won 4--1~\cite{gibney2016google,hassabis2016alphago}. A system that was stronger than its opponent overall still failed after an unexpected move.

This study examines large language models (LLMs) used as chess evaluators. Gradient-based attacks such as the fast gradient sign method (FGSM)~\cite{goodfellow2015explaining} and projected gradient descent (PGD)~\cite{madry2018towards} need the gradients of the model, which commercial LLM interfaces do not expose. We therefore perturb the input in a different way: we apply transformations to a chess position that leave its game-theoretic value unchanged, or change it in a known way, and test whether the model's evaluation changes accordingly. The approach needs only query access to the model.

Our aim is to measure how consistently LLMs evaluate positions that a correct evaluator would treat as equivalent, and to compare this consistency with the models' accuracy against an engine. The same question, whether a model gives consistent outputs for inputs that should be treated alike, arises when AI systems are used in higher-stakes domains such as autonomous driving and medical diagnosis. The main contributions of this paper are as follows:
\begin{itemize}
\item We apply to LLM evaluators three transformations that earlier work used to test chess engines: a $180^\circ$ rotation of pawnless positions, left--right mirroring and color swapping~\cite{mendez2023metamorphic,fluri2023evaluating}. We add tests on position pairs one move apart, on conversion between Forsyth--Edwards Notation (FEN) and a natural-language description, and on illegal positions (Section~\ref{sec:methods}).
\item We apply them to six LLMs accessed through their application programming interfaces (APIs) and find that all six fail on rotated positions: at most 10.6\% of the 500 rotated pairs receive evaluations within 300 centipawns (cp) of each other, and most \texttt{null} answers to rotated positions wrongly declare them illegal (Section~\ref{sec:geom}).
\item We classify the outcome of every pair, including refusals, failed calls and mate-level scores, and show that a mean computed only over pairs with two ordinary scores rests on 32 to 366 of the 500 rotated pairs, depending on the model, and ranks the models misleadingly (Sections~\ref{sec:stability} and~\ref{sec:tradeoff}).
\item On the other four transformations, the ranking of the models by stability matches their ranking by accuracy against Stockfish except for the first two places, and pairs whose evaluations differ by more than 300~cp are common for every model (Sections~\ref{sec:tradeoff} and~\ref{sec:errors}).
\end{itemize}

\section{Background and Related Work}\label{sec:background}

\subsection{State Representation}

FEN~\cite{edwards1994pgn} encodes a single position as a compact string: the piece placement rank by rank, the side to move, the castling rights, the en passant target square, the halfmove clock and the fullmove number (e.g., \texttt{rnbqkbnr/pppppppp/8/8/8/8/PPPPPPPP/RNBQKBNR w KQkq - 0 1}). It is designed for exact, machine-readable snapshots. Portable Game Notation (PGN)~\cite{edwards1994pgn} records an entire game: a header of tagged metadata (event, site, players, ratings and so on) and a movetext section in Standard Algebraic Notation (SAN), optionally with comments and the result (e.g., \texttt{1.e4 e5 2.Nf3 Nc6 ... 1-0}). Natural-language descriptions use free-form prose for positions or moves (e.g., ``White sacrifices the exchange on move 23 and wins the endgame by pushing the passed a-pawn''), favoring human narrative over strict form.

FEN serializes states: it is precise, minimal and suited to engine input and output, databases and reproducible diagrams, but it carries no history. PGN serializes processes: it preserves move order, annotations and metadata for training, research and archiving, but reconstructing a middlegame position requires replaying the moves. Natural language serves interpretation: it conveys plans, themes and teaching points and tolerates differences in style, but it is ambiguous, lossy and not directly executable by software. In practice, FEN underlies position-level tasks, PGN underlies game-level analysis, and natural language adds explanatory context that neither structured format captures. We use FEN and a natural-language description generated from it by a template (Section~\ref{sec:format}); both are established forms of input, which keeps the evaluation zero-shot and avoids non-standard representations~\cite{jhamtani2018learning,fiekas_pythonchess}.

\subsection{Value Function}

A centipawn (cp)~\cite{campbell2002deep,guid2006computer} is one hundredth of a pawn, so 1~pawn = 100~cp. Engine evaluations are real-valued scores in centipawns, reported relative to a fixed side by convention: many interfaces use White's perspective, and some tools report from the side to move. Thus $+100$~cp means that the reference side is about a pawn better, 0~cp is roughly equal, and $-50$~cp is a disadvantage of half a pawn.

Centipawn scores combine material, king safety, piece activity, pawn structure and other features into a single number that engines optimize during search. A centipawn score is not a probability, but engines often map it to an expected score through an engine-specific, approximately logistic calibration. Larger absolute values indicate a more decisive advantage.

\subsection{Recommended Moves}

A recommended move is the move that an engine or a player considers best in a position. Engines base the recommendation on an evaluation function that considers piece activity, material balance, king safety, pawn structure and tactical threats; they search millions of move sequences and return the best move or a set of strong candidates. The evaluation assigns each candidate move a value for the position after the move, and a higher value favors the side to move.

Engines compute these recommendations with different algorithms. Stockfish searches with alpha--beta pruning to a set depth and evaluates positions with an efficiently updatable neural network (NNUE)~\cite{nasu2018nnue,stockfish_nnue_blog}; since version 16.1 it no longer uses a hand-crafted evaluation~\cite{stockfish_16_1}. AlphaZero~\cite{silver2018general} and Leela Chess Zero~\cite{lc0_releases} learn a policy and value network through self-play and select moves with Monte Carlo tree search (MCTS), relying on learned patterns rather than predefined rules. This difference in methods reflects a shift in chess engines from hand-crafted rules to learned evaluation.

\subsection{Large Language Models}

Widely used LLM families include the proprietary GPT~\cite{achiam2023gpt4,bang2023multitask}, Claude~\cite{schmidl2024assessing} and Gemini~\cite{gemini2023} series and the open-weight DeepSeek~\cite{liu2024deepseek}, Llama~\cite{dubey2024llama}, Qwen~\cite{yang2025qwen3}, Mistral~\cite{jiang2023mistral} and Gemma~\cite{team2025gemma} models. Training for reasoning has also advanced: DeepSeek-R1 showed that the reasoning abilities of LLMs can be incentivized through reinforcement learning without human-labelled reasoning trajectories~\cite{guo2025deepseekr1}.

\subsection{Chess Engines}

Chess engines are standard tools for analyzing games and finding strong moves. Widely used engines include Stockfish~\cite{stockfish_nnue_blog,stockfish_17_1,nasu2018nnue}, Leela Chess Zero (Lc0)~\cite{lc0_releases} and Komodo~\cite{haworth2021tcec20}. Stockfish is one of the strongest open-source engines. AlphaZero~\cite{silver2017mastering,silver2018general}, a research system, and Lc0 are trained by self-play and select moves with MCTS.

\subsection{Chess-Specific Evaluations of LLMs}

Schultz et~al.~\cite{schultz2025mastering} showed that combining a language model with search, either by guiding external MCTS or by generating a linearized search tree in context, improves playing strength in chess, Chess960, Connect Four and Hex, and reaches grandmaster-level performance in chess with a search budget closer to that of human players. Wen et~al.~\cite{wen2025chessqa} introduced ChessQA, a dynamic and reproducible benchmark that tests LLM chess understanding beyond move quality in five categories (structural understanding, motifs, short tactics, position judgment and semantic description); the models they evaluated showed persistent weaknesses in all five. Our stability framework complements such accuracy-driven benchmarks: it tests whether evaluations stay consistent under transformations that preserve game-theoretic equivalence.

Kumar and Mishra~\cite{kumar2025robustness} argue that the robustness of LLMs has to be assessed beyond conventional accuracy criteria. Surveys of hallucination in natural language generation~\cite{ji2023survey} and of the evaluation of large multimodal models~\cite{zhang2025lmm} describe further failure modes and evaluation dimensions. LLMs also struggle with text-based spatial reasoning, where coupling an LLM with a symbolic solver improves accuracy~\cite{wang2024dspy}.

\subsection{Metamorphic Testing}

Metamorphic testing (MT) addresses the oracle problem, in which checking the correctness of an individual output is difficult or expensive~\cite{chen1998metamorphic}. Instead of checking a single output, MT checks metamorphic relations (MRs): properties that must hold between the outputs for related inputs. We define MRs from the geometric and game-theoretic symmetries of chess. A transformation such as a board rotation or a color swap turns a source input $S$ into a follow-up input $S'$. The token sequence changes, but the game-theoretic content is preserved: the value of the position is unchanged or, for a color swap, negated. A model whose evaluation depends on the position rather than on how it is written should return the corresponding value for $S'$. A violation of an MR therefore shows that the output depends on the encoding of the position and not only on the position itself. Chan et~al.~\cite{chan2025symmetric} used symmetric MRs to test the stability of code-generation LLMs under equivalent rewritings of the prompt, which supports the use of MRs as a test of LLM stability.

Metamorphic relations and similar consistency checks have been applied to chess engines. M\'{e}ndez et~al.~\cite{mendez2023metamorphic} defined MRs for Stockfish, among them symmetric versions of a position and a color swap, which should leave the evaluation unchanged. On a dataset of more than 40,000 positions, they found non-negligible deviations from the expected behavior for every relation. Martin et~al.~\cite{martin2025reevaluating} reproduced these deviations under the original conditions, which included a search depth of 10 plies. With deeper searches, positions from real games and different versions of Stockfish, they found the engine much more consistent. They also report that for each position there is a depth beyond which the relations no longer produce differences in evaluation. Fluri et~al.~\cite{fluri2023evaluating} proposed consistency checks for models whose individual decisions are hard to verify and applied them to Leela Chess Zero and Stockfish. Their four checks concern forced moves, rotations and reflections of pawnless positions without castling rights, swaps of the two sides' pieces, and the model's own recommended moves. The value of the position should not change in the first three cases and should change little after a recommended move, and both engines violated all four checks. We apply the rotation, mirror and color-swap relations to LLMs and add two tests aimed at them: conversion between FEN and a natural-language description, and illegal positions.

\section{Materials and Methods}\label{sec:methods}

Sections~\ref{sec:pairs}--\ref{sec:forced} define the seven test sets, and Section~\ref{sec:setup} describes the models, the reference engine, the data and the metrics. The prompts are listed in Appendix~\ref{app:prompts}.

\subsection{Position Pairs One Move Apart}\label{sec:pairs}

Each pair consists of a base position $A$ and the position $B$ reached from $A$ by one legal move chosen at random. The base positions come from Lichess games of the Atomic variant (Section~\ref{sec:data}) and were taken after 20 to 24 plies, according to their FEN move counters. Both positions of every evaluated pair are legal in standard chess according to python-chess, and the move from $A$ to $B$ is legal under the standard rules. A candidate pair is accepted only if it passes two filters (Algorithm~\ref{alg:pairs}). First, the piece-placement fields of the two FEN strings must have a similarity ratio of at least 0.9, computed with the \texttt{SequenceMatcher} class of Python's \texttt{difflib} module. Second, Stockfish evaluates both positions to depth 10, and the two evaluations from White's perspective must differ by at least 80~cp. The procedure therefore selects pairs whose FEN strings are nearly identical but whose evaluations differ. Mate scores were not filtered out: 5 of the 500 evaluated pairs contain one, and these pairs are excluded from Tables~\ref{tab:stability} and~\ref{tab:accuracy} (Section~\ref{sec:metrics}).

\begin{algorithm}[H]
\caption{Sampling position pairs with high FEN similarity and a large difference in evaluation. The thresholds were $\mathit{SimilarityThreshold} = 0.9$ and $\mathit{CpDifferenceThreshold} = 80$~cp, with $\mathit{TargetNumber} = 5000$ and $\mathit{MaxTrials} = 500{,}000$.}\label{alg:pairs}
\begin{algorithmic}[1]
\State Initialize $\mathit{AcceptedPairs} \gets \emptyset$, $\mathit{Trials} \gets 0$
\While{$\mathrm{Count}(\mathit{AcceptedPairs}) < \mathit{TargetNumber} \land \mathit{Trials} < \mathit{MaxTrials}$}
  \State $\mathit{Trials} \gets \mathit{Trials} + 1$
  \State $\mathit{board\_A} \gets$ a base position drawn at random
  \If{$\mathit{board\_A}$ has no legal move}
    \State \textbf{continue}
  \EndIf
  \State $\mathit{board\_B} \gets \mathit{board\_A}$ after one legal move drawn at random
  \State $\mathit{similarity} \gets \mathrm{PiecePlacementSimilarity}(\mathit{board\_A}, \mathit{board\_B})$
  \If{$\mathit{similarity} < \mathit{SimilarityThreshold}$}
    \State \textbf{continue}
  \EndIf
  \State $\mathit{cp\_A} \gets \mathrm{GetWhiteEngineScore}(\mathit{board\_A})$
  \State $\mathit{cp\_B} \gets \mathrm{GetWhiteEngineScore}(\mathit{board\_B})$
  \If{$\mathit{cp\_A}$ is \texttt{null} $\lor$ $\mathit{cp\_B}$ is \texttt{null}}
    \State \textbf{continue}
  \EndIf
  \If{$|\mathit{cp\_B} - \mathit{cp\_A}| \ge \mathit{CpDifferenceThreshold}$}
    \State Add $(\mathit{board\_A}, \mathit{board\_B})$ to $\mathit{AcceptedPairs}$
  \EndIf
\EndWhile
\State \Return $\mathit{AcceptedPairs}$
\end{algorithmic}
\end{algorithm}

\subsection{Rotation Test on Pawnless Positions}\label{sec:rotation}

To test whether an LLM's evaluation is invariant under a rotation of the board, rather than tied to the usual placement of the pieces, we use pawnless positions. A $180^\circ$ rotation maps each square to the opposite square through the center of the board (a1 to h8, b1 to g8, a2 to h7 and so on) and keeps the colors of the pieces and the side to move. Without pawns, whose direction of movement depends on their color, and without castling rights, the rotated position has the same game-theoretic value as the original. Rotations by $90^\circ$ and $270^\circ$ were not tested.

The positions were constructed in two steps. A random number of random legal moves was played from the initial position, and all pawns were then removed from the resulting position; positions that python-chess~\cite{fiekas_pythonchess} reported as invalid after the removal were discarded. These positions are legal but were not reached by play, and their move counters (3 to 120 plies) refer to the game before the pawns were removed. In 223 of the 500 evaluated pairs, the original position has castling rights, which python-chess removes from the rotated position because its kings and rooks no longer stand on their initial squares. A pair was kept only if Stockfish's evaluations of the two positions at depth 10 differed by at most 80~cp and the win probabilities derived from Stockfish's win--draw--loss estimates differed by at most 0.05. At this depth, Stockfish itself can evaluate equivalent positions differently~\cite{mendez2023metamorphic,martin2025reevaluating}; the filter removed the pairs on which its two evaluations differed by more than these tolerances. Positions with a mate score or with an evaluation above 4000~cp in magnitude were excluded. For the 223 pairs with castling rights, the two positions are equivalent only within this tolerance. The LLM evaluates both positions with the same prompt, and we compare the two values from White's perspective. Because rotation preserves the game-theoretic value, a model that encodes the position consistently should return the same value for both.

When the engine scores the two positions within this tolerance, large differences between the LLM's evaluations of them indicate a dependence on coordinate patterns rather than on a stable representation of the position. The test is cheap to run and gives measurable targets for remedies such as symmetry-aware prompting, training data augmented with rotated and reflected boards, or structured state encodings.

\subsection{Mirror Test (Left--Right Reflection)}\label{sec:mirror}

We test whether an LLM's evaluation is invariant under a left--right reflection of the board, which maps file a to h, b to g and so on, and keeps the ranks and the side to move. In standard chess this reflection preserves the rules only when neither side has castling rights, because in the reflected position the kings stand on the d-file, from which standard castling is not possible. We computed the reflection with python-chess, which removes the castling rights that the reflected position cannot have; this applied to 349 of the 500 evaluated pairs. The base positions were reached by random legal moves from the initial position (1 to 80 plies). As for rotation, a pair was kept only if Stockfish's evaluations at depth 10 differed by at most 80~cp and the win probabilities by at most 0.05. Positions with a mate score or an evaluation above 4000~cp in magnitude were excluded. For the 349 pairs whose original has castling rights, the two positions are equivalent only within this tolerance. For the other pairs, the mirror has the same game-theoretic value as the original, and a consistent evaluator should return the same centipawn value for both.

The protocol is as follows. (i) Select a legal position and construct its mirror by applying the file permutation $a \leftrightarrow h, b \leftrightarrow g, \ldots$ to the pieces and to the en passant square. (ii) Keep the move number, the side to move and the clocks. (iii) Query the LLM with identical prompts for the original and the mirrored position. (iv) Compare the two values. A sign flip between them would indicate a preference induced by the coordinates rather than by the position.

Keeping castling rights that the mirrored position cannot have would change which moves are legal and make the pair non-equivalent. Removing them, as here, avoids this, but when castling was still possible in the original, the two positions differ in their legal moves and are equivalent only within the engine's tolerance.

Consistency across mirrored positions is a minimal test of geometric consistency. When the engine confirms equivalence but the LLM gives very different evaluations, or reasons asymmetrically about kingside and queenside play, the model is likely to rely on coordinate templates or lexical priors. The test is cheap and gives a measurable target for remedies such as training data augmented with reflected boards, symmetry-aware encodings or post hoc repair under constraints.

\subsection{Color-Swap Test}\label{sec:colorswap}

We test whether an LLM's evaluation respects the color symmetry of chess. Given a legal position $P$, the color-swapped position $T_{\mathrm{col}}(P)$ is obtained by reflecting the board across its horizontal midline (rank 1 to rank 8, rank 2 to rank 7 and so on), exchanging the colors of all pieces, toggling the side to move, exchanging the castling rights of the two sides ($KQ \leftrightarrow kq$) and reflecting the en passant square (e.g., e3 to e6). The clocks and the move number are kept. We constructed $T_{\mathrm{col}}(P)$ with python-chess's \texttt{Board.mirror()}, which performs exactly these operations. Because the rules treat the two colors symmetrically, $T_{\mathrm{col}}(P)$ has the same game-theoretic value as $P$ with the roles of the two sides exchanged. With evaluations reported from a fixed perspective (White's), this gives
\begin{equation}
E(P) \approx -E\bigl(T_{\mathrm{col}}(P)\bigr),
\label{eq:antisym}
\end{equation}
so the two scores should have equal magnitude and opposite sign.

The protocol is as follows. (1) Generate positions by random legal moves from the initial position (1 to 79 plies). Keep a position only if Stockfish's evaluations at depth 10 satisfy $|E_{\mathrm{SF}}(P) + E_{\mathrm{SF}}(T_{\mathrm{col}}(P))| \le 60$~cp, and exclude positions with a mate score or an evaluation above 4000~cp in magnitude. (2) Construct $T_{\mathrm{col}}(P)$. (3) Query the LLM with identical prompts for $P$ and $T_{\mathrm{col}}(P)$, asking for centipawn values from White's perspective. (4) Measure the anti-symmetry gap
\begin{equation}
\Delta_{\pm} = \bigl|E(P) + E\bigl(T_{\mathrm{col}}(P)\bigr)\bigr|,
\label{eq:gaps}
\end{equation}
which is zero for a model that satisfies Equation~(\ref{eq:antisym}) exactly.

For example, in a middlegame position with White to move, Stockfish gives $E_{\mathrm{SF}}(P) = +25$~cp, a slight edge for White. The color-swapped position has Black to move, and Stockfish gives $E_{\mathrm{SF}}(T_{\mathrm{col}}(P)) = -25$~cp from White's perspective, so $\Delta_{\pm} = 0$. This is the reference behavior for the LLM.

A common error is to mix perspectives: some systems report scores from the side to move, others from White's side, and the test is valid only under a fixed perspective. When the engine shows near-perfect anti-symmetry but the LLM does not, the model is likely to rely on color-specific lexical priors or coordinate templates rather than on a representation that is invariant to the side. The color-swap test thus complements the geometric tests of rotation and reflection.

\subsection{Cross-Format Consistency}\label{sec:format}

We test whether an LLM gives consistent evaluations of the same position in two representations: (i) its FEN string and (ii) a natural-language description generated from the FEN by a template. The description states the side to move, lists the square of every piece by color and type, and gives the castling rights, any en passant target square, the halfmove clock and the fullmove number (Appendix~\ref{app:nlexample}). It contains all the information in the FEN, in a different form. The positions were taken from rated standard games on Lichess (Section~\ref{sec:data}) after 8 to 20 plies.

For each position we query the LLM twice, once per input, with the templates of Appendix~\ref{app:prompts} and otherwise identical settings, and record the format gap $|E_{\mathrm{FEN}} - E_{\mathrm{NL}}|$ between the two evaluations. Stockfish does not grade the natural-language input; its evaluation of the FEN serves as the accuracy reference (Section~\ref{sec:metrics}).

Because the description restates the FEN without loss, a difference between the two evaluations cannot come from missing information. It comes from the surface form of the input or from variation between repeated calls, which a single evaluation per input cannot separate (Section~\ref{sec:limitations}). Systematic deviations of this kind motivate schema-guided descriptions or training with cross-format consistency losses.

\subsection{Illegal-Position Test}\label{sec:illegal}

We test whether an LLM detects positions that violate the rules of chess and declines to evaluate them. Unlike the transformations above, these inputs corrupt the state itself. The prompt (Appendix~\ref{app:prompts}) instructs the model to check the position and, if it is illegal, to set both numerical fields to \texttt{null} and to describe the violation in the \texttt{error} field.

The positions were generated by a program that places pieces according to one of six patterns: two White kings; no Black king; a White pawn on the first rank; ranks 2 to 7 filled with pawns, which gives each side more than eight pawns; two kings and two rooks on random squares, labelled as both kings in check; and a pawn of each color on its starting rank, labelled as an invalid en passant square. The other FEN fields were drawn at random: the side to move, the castling rights, an en passant square in about 30\% of the positions, and the clocks. All 500 evaluated positions fail python-chess's validity check, 369 of them for more than one reason. In the last two patterns the labelled violation is often absent: the side not to move is in check in only 25 of the 81 positions of the first, and the en passant square is invalid in only 33 of the 67 positions of the second. Of the other 90 positions, 88 have castling rights or an en passant square that the placement of the pieces does not allow, and 7 lack a king.

We count an answer as a rejection if \texttt{centipawn\_white} is \texttt{null}, as the prompt requires. Any numerical evaluation of an illegal position counts as a failure, even if the \texttt{error} field names a violation; this occurred once (Claude Sonnet 4.5). Calls that returned no parsable answer are reported separately (Section~\ref{sec:models}). The prompt does not ask for a move, so move recommendations are not scored.

Illegal positions test whether a model represents the constraints of chess and not only its statistical surface. Reliable systems should validate the input before evaluating it, for example with FEN sanity checks and a test of whether the position can arise from a legal game. Rejection rates in our experiments ranged from 44.2\% to 96.0\% (Table~\ref{tab:safety}), which suggests a role for rule-based tools (schema validation, cross-checks with a move generator) or for training objectives that penalize the scoring of positions that cannot arise in a legal game.

\subsection{Forced-Move Positions}\label{sec:forced}

This set contains positions from standard and Atomic games on Lichess (Section~\ref{sec:data}) in which, under the standard rules, the side to move has exactly one legal move; in 479 of the 500 evaluated positions, the side to move is in check. Stockfish evaluated each position to depth 12, and positions with a mate score or with an evaluation above 4000~cp in magnitude were excluded when the set was built. No evaluated position has a forced mate that Stockfish found at this depth, and the largest absolute evaluation among the 500 is 1276~cp. The set tests whether a model's evaluation of a forcing position agrees with the engine; we report the accuracy error and the number of mate-level scores (Section~\ref{sec:metrics}).

\subsection{Experimental Setup}\label{sec:setup}

\subsubsection{Computing Environment}

The experiments ran on an Intel Core Ultra 5 125H processor (14 cores and 18 threads) with 15~GB of memory, under Ubuntu 22.04.5 LTS in the Windows Subsystem for Linux 2 (WSL2) and with Python 3.10.12. No GPU was used: the LLMs were queried through their providers' APIs, and Stockfish and the analysis scripts ran on the CPU.

\subsubsection{Models}\label{sec:models}

We evaluated six LLMs, identified below by their API names. The six models are the objects of this study, and their outputs are the data we analyze.
\begin{itemize}
\item GPT-5.1 (\texttt{gpt-5.1}, OpenAI). The API default for GPT-5.1 is no reasoning (reasoning effort \texttt{none})~\cite{openai_gpt51}, and the requests did not change it.
\item Claude Sonnet 4.5 (\texttt{claude-sonnet-4-5-20250929}, Anthropic). Extended thinking is used only when the request enables it~\cite{anthropic_thinking}, and the requests did not enable it.
\item Gemini 2.5 Flash (\texttt{gemini-2.5-flash}, Google). Thinking is on by default for this model, with the amount of reasoning set dynamically~\cite{google_thinking}. The requests did not configure it, so the model reasoned before answering. In 144 calls the API returned no text and reported the finish reason \texttt{MAX\_TOKENS}, which means that the token limit set in the request was reached~\cite{google_generate}.
\item DeepSeek Chat (\texttt{deepseek-chat}, DeepSeek). This name refers to the non-thinking mode of DeepSeek's current V3-series model; it pointed to DeepSeek-V3.2-Exp from 29 September 2025 and to DeepSeek-V3.2 from 1 December 2025~\cite{deepseek_updates}. All DeepSeek outputs had been committed to the project repository by 30 November 2025 and therefore come from DeepSeek-V3.2-Exp.
\item Kimi K2 Turbo (\texttt{kimi-k2-turbo-preview}, Moonshot AI), a high-speed version of Kimi K2~\cite{kimi_quickstart}.
\item Grok 4.1 Fast (\texttt{grok-4-1-fast-non-reasoning}, xAI), the non-reasoning variant~\cite{xai_grok}.
\end{itemize}

All requests used the same settings: temperature 0, at most 8000 output tokens, the provider's default nucleus sampling (top-$p$ was set to 1.0 explicitly only for Gemini 2.5 Flash), a JSON response format where the API offers one, and a timeout of 120~s. JSON mode was requested from the OpenAI, DeepSeek, Moonshot and xAI interfaces and the JSON MIME type from the Gemini interface; for Claude Sonnet 4.5, the output schema was given in the system message. The prompt template of Appendix~\ref{app:prompts} was sent as the user message, after a system message that asked for engine-like evaluations in JSON. Each input was evaluated once. A request was repeated up to three times after an API error or an answer that could not be parsed as a JSON object with the two numerical fields. An input without a parsed answer after four attempts counts as a failed call. GPT-5.1, DeepSeek Chat and Grok 4.1 Fast had no failed calls, Claude Sonnet 4.5 had 3 (unparsable output), Gemini 2.5 Flash had 144 (the empty responses above) and Kimi K2 Turbo had 455 (452 rate-limit errors and 3 timeouts). The outputs were committed to the project repository between 28 November and 1 December 2025.

In preliminary tests, the reasoning variant of Grok 4.1 Fast used 33,107 reasoning tokens to produce an answer of 123 tokens, so we used the non-reasoning variant to keep cost and latency comparable across models. Gemini 2.5 Flash, by contrast, reasoned by default, so the comparison mixes one reasoning model with five non-reasoning ones (Section~\ref{sec:limitations}).

\subsubsection{Reference Engine}\label{sec:engine}

We used Stockfish 16.1~\cite{stockfish_16_1} through the python-chess library~\cite{fiekas_pythonchess} (version 1.11.2), with 2 search threads and a 128~MB hash table. The test sets record Stockfish's evaluation of every position at depth 10 (depth 12 for the forced-move set), and these values are the reference for the accuracy error. Positions with a mate score were excluded from the color-swap, rotation, mirror and forced-move sets when the sets were built. In the other sets, 5 of the 500 evaluated similarity pairs and 1 of the 500 format positions have a mate score, and they are excluded from Tables~\ref{tab:stability} and~\ref{tab:accuracy} by the rule of Section~\ref{sec:metrics}.

\subsubsection{Data Sources}\label{sec:data}

Each of the seven test sets was generated with 5000 items, and the models evaluated the first 500 items of each set. Each item of the five paired sets requires two queries and each forced-move or illegal position one, which gives 3500 items and 6000 queries per model.

Lichess games. The similarity, format-conversion and forced-move sets use positions from games in the October 2024 files of the Lichess open database~\cite{lichess_db}, which publishes one PGN file per month for standard chess and for each variant. The format-conversion positions come from rated standard games, the base positions of the similarity set from rated games of the Atomic variant, and the forced-move positions from both. Atomic has different capture rules, so a position reached in an Atomic game need not be reachable in standard chess; every evaluated position from these sets is nevertheless legal in standard chess according to python-chess.

Random play. To reduce bias from data contamination, that is, from a model having memorized well-known games, the color-swap, mirror and rotation sets use positions reached by random legal moves from the initial position; for rotation, the pawns were then removed (Section~\ref{sec:rotation}). Such positions are unlikely to appear in published games.

Synthetic positions. The illegal positions were generated by a program (Section~\ref{sec:illegal}).

\subsubsection{Metrics}\label{sec:metrics}

Each query has one of three outcomes: a numerical evaluation, a \texttt{null} evaluation (an abstention, which the prompt requires for illegal positions and allows for legal positions on which the model is uncertain) or a failed call. A numerical evaluation above 10,000~cp in magnitude is a mate-level score: the prompt gives $-10{,}000$ to $+10{,}000$ as the typical range and asks for values such as $\pm$30,000 for mates (Appendix~\ref{app:prompts}). We use the following metrics.
\begin{itemize}
\item Stability error (Table~\ref{tab:stability}): the mean, over pairs, of the gap between the two evaluations of a pair. The gap is $\Delta_{\pm}$ (Equation~(\ref{eq:gaps})) for the color swap, $|E(P) - E(P')|$ for rotation and mirroring, $|E_{\mathrm{FEN}} - E_{\mathrm{NL}}|$ for format conversion, and $\bigl||E(A) - E(B)| - |E_{\mathrm{SF}}(A) - E_{\mathrm{SF}}(B)|\bigr|$ for the pairs one move apart; the last compares the size of the change in the model's evaluation with the size of the change in Stockfish's and ignores its direction. A pair enters the mean only if both evaluations are numerical and at most 10,000~cp in magnitude (for the pairs one move apart, both Stockfish evaluations as well). Table~\ref{tab:stability} gives the number of such pairs.
\item Outcome of a pair (Figure~\ref{fig:outcomes}): this applies to the four transformations that should preserve or negate the value (color swap, rotation, mirror and format conversion). A pair that enters Table~\ref{tab:stability} is consistent if its gap is at most 300~cp and inconsistent otherwise. A pair whose two evaluations are numerical but include a mate-level score is mate-level, and a pair with at least one \texttt{null} evaluation or failed call is unanswered. The consistent share is the number of consistent pairs divided by 500. The threshold of 300~cp is our convention, chosen to be roughly the value of a minor piece, so that a larger gap indicates a disagreement about the state of the game.
\item Accuracy error (Table~\ref{tab:accuracy}): the mean absolute difference between the LLM's evaluation and Stockfish's, $\mathrm{mean}(|cp_{\mathrm{LLM}} - cp_{\mathrm{SF}}|)$, over the original positions of the color-swap, rotation, mirror and format-conversion sets (FEN input) and over both positions of each pair one move apart, with the same exclusion rule as above. Table~\ref{tab:accuracy} reports these five task-level values and their mean. For the forced-move set, the accuracy error is computed over the answers that are not mate-level scores (Table~\ref{tab:errors}).
\item Rejection rate (Table~\ref{tab:safety}): the percentage of the 500 illegal positions for which the model returned a \texttt{null} evaluation.
\end{itemize}

\subsubsection{Use of Generative AI}\label{sec:genai}

Apart from the six models under study, we used Claude (Anthropic; Claude Opus 5.5) in preparing this version of the article: to check the analysis code against the raw model outputs; to recompute Tables~\ref{tab:stability}--\ref{tab:safety} and to compute Tables~\ref{tab:errors} and~\ref{tab:outcomes} and Figures~\ref{fig:tradeoff} and~\ref{fig:outcomes} with scripts that only read the archived data; to draw the figures; to search for related work and check the references; and to draft and edit the text. The scripts and their outputs are included in the Zenodo archive (Data Availability Statement).

\section{Results}\label{sec:results}

We report results in five areas: stability under five types of transformation (Section~\ref{sec:stability}), accuracy against Stockfish 16.1 at depth 10 (Section~\ref{sec:accuracy}), the relation between the two (Section~\ref{sec:tradeoff}), the detection of illegal positions (Section~\ref{sec:safety}) and the counts of inconsistent pairs and mate-level scores (Section~\ref{sec:errors}). Our framework gives stability priority over raw accuracy, on the premise that a reliable evaluator must satisfy the geometric and semantic consistency relations of Section~\ref{sec:methods}.

\subsection{Stability}\label{sec:stability}

Stability is measured as the mean gap, in centipawns, between the evaluations of a position and its transformed counterpart (Section~\ref{sec:metrics}). An ideal model satisfies $V(P) \approx V(P')$, or $V(P) \approx -V(P')$ for the color swap, and has an error near zero. Table~\ref{tab:stability} gives the results and the number of pairs that enter each mean. A pair enters only if both evaluations are numerical and at most 10,000~cp in magnitude, and this condition excludes different numbers of pairs for different models and transformations. For rotation, the means rest on 32 (DeepSeek Chat) to 366 (GPT-5.1) of the 500 pairs; for the other four transformations, on 274 to 496 pairs. The rotation means are therefore not comparable across models (Section~\ref{sec:geom}). Table~\ref{tab:stability} also gives the mean over the other four columns, which ranges from 186.17~cp (GPT-5.1) to 636.61~cp (Grok 4.1 Fast).

\begin{table}[htbp]
\footnotesize\setlength{\tabcolsep}{3pt}
\caption{Stability errors: mean gap (cp) between the evaluations of paired positions for each transformation, and the number of pairs that enter each mean; lower errors are better. A pair enters only if both evaluations are numerical and at most 10,000~cp in magnitude (Section~\ref{sec:metrics}). The rotation means rest on very different numbers of pairs and are not comparable across models (Section~\ref{sec:geom}). Avg.\ is the unweighted mean of the five columns, and Avg.\ w/o Rot.\ that of the four columns other than rotation. Bold marks the lowest error in each column except Rotation and Avg.\label{tab:stability}}
\begin{tabularx}{\linewidth}{lCCCCCCC}
\toprule
\textbf{Model} & \textbf{Color Swap} & \textbf{Rotation} & \textbf{Mirror} & \textbf{Similarity} & \textbf{Format} & \textbf{Avg.} & \textbf{Avg.\ w/o Rot.} \\
\midrule
\multicolumn{8}{l}{\textit{Mean gap (cp)}} \\
GPT-5.1           & 217.49 & 2531.28 & 181.08 & \textbf{297.99} & 48.11 & 655.19 & \textbf{186.17} \\
Claude Sonnet 4.5 & \textbf{198.78} & 626.56 & 195.85 & 318.25 & 62.41 & 280.37 & 193.82 \\
Kimi K2 Turbo     & 367.67 & 726.21 & \textbf{94.94} & 329.38 & \textbf{36.22} & 310.88 & 207.05 \\
DeepSeek Chat     & 254.46 & 996.88 & 235.18 & 399.58 & 121.48 & 401.52 & 252.68 \\
Gemini 2.5 Flash  & 776.80 & 854.60 & 747.48 & 559.34 & 260.56 & 639.75 & 586.04 \\
Grok 4.1 Fast     & 785.95 & 1046.98 & 516.88 & 767.25 & 476.37 & 718.69 & 636.61 \\
\midrule
\multicolumn{8}{l}{\textit{Pairs entering the mean (of 500)}} \\
GPT-5.1           & 482 & 366 & 486 & 485 & 491 & & \\
Claude Sonnet 4.5 & 459 & 48  & 401 & 457 & 491 & & \\
Kimi K2 Turbo     & 422 & 87  & 390 & 429 & 441 & & \\
DeepSeek Chat     & 471 & 32  & 469 & 478 & 496 & & \\
Gemini 2.5 Flash  & 298 & 50  & 274 & 319 & 414 & & \\
Grok 4.1 Fast     & 360 & 104 & 354 & 421 & 395 & & \\
\bottomrule
\end{tabularx}
\end{table}

\subsubsection{Rotation and Mirror}\label{sec:geom}

\begin{figure}[htbp]
\centering
\includegraphics[width=\textwidth]{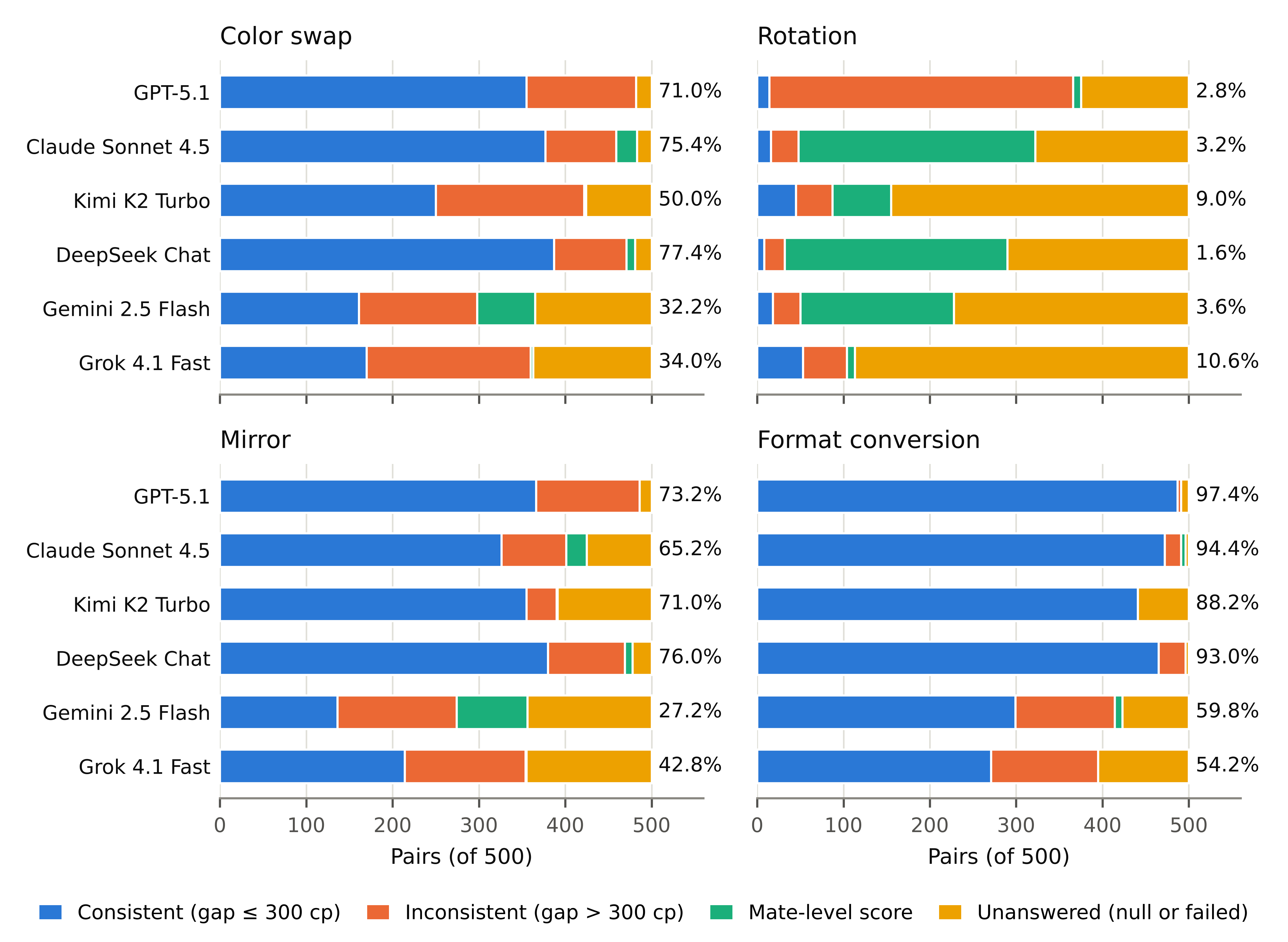}
\caption{Outcome of the 500 pairs of each transformation for each model. Consistent: both evaluations numerical and at most 10,000~cp in magnitude, with a gap of at most 300~cp; inconsistent: the same with a gap above 300~cp; mate-level: at least one score above 10,000~cp in magnitude; unanswered: at least one evaluation \texttt{null} or a failed call. The number at the right of each bar is the consistent share. Table~\ref{tab:outcomes} lists the counts.\label{fig:outcomes}}
\end{figure}

Rotation gave the highest mean error of the five transformations for every model (Table~\ref{tab:stability}), but these means describe only the pairs in which both evaluations were ordinary scores. Figure~\ref{fig:outcomes} classifies all 500 pairs of each transformation. For rotation, the consistent share ranged from 1.6\% (DeepSeek Chat) to 10.6\% (Grok 4.1 Fast), so for every model at most about one rotated pair in ten received evaluations within 300~cp of each other. The other pairs failed in different ways for different models. GPT-5.1 gave ordinary scores to both positions in 366 pairs, but the two scores differed by more than 300~cp in 352 of them, and these failures remain in its mean. Claude Sonnet 4.5, DeepSeek Chat and Gemini 2.5 Flash answered both positions numerically but gave a mate-level score for at least one of them in 274, 258 and 178 pairs, and Kimi K2 Turbo and Grok 4.1 Fast left at least one position unanswered in 345 and 387 pairs; these pairs do not enter Table~\ref{tab:stability}. As a result, GPT-5.1's rotation mean includes most of its failures, whereas the other models' rotation means rest on the few pairs they answered with ordinary scores. The largest rotation error, that of GPT-5.1 (2531.28~cp), does not show that GPT-5.1 is less stable under rotation than the other models.

Most \texttt{null} answers to rotated positions declared a legal position illegal. Across the six models, 1367 rotated positions received a \texttt{null} evaluation, and in 1324 of these answers the \texttt{error} field described the position as illegal or impossible. In 818 of them, the stated reason was that one side had more than one king (e.g., ``Illegal position: White has two kings (on d1 and e1)'', from Claude Sonnet 4.5), although every rotated position has exactly one king of each color. The same 500 positions before rotation received 355 \texttt{null} evaluations. These error messages describe pieces that are not on the board, which indicates that the models misread the rotated FEN strings; in 444 of the 500 rotated positions, the White king stands on the seventh or eighth rank.

One possible reason why rotation is the hardest transformation is that it changes the FEN string more than mirroring does. A $180^\circ$ rotation reverses both the order of the ranks and the order of the squares within each rank, whereas a mirror only reverses the order within each rank and leaves each rank in its place. The rotation set also differs from the others in its positions: they are pawnless, were constructed by removing the pawns from positions reached by random play, and include 223 pairs whose original has castling rights that the rotated position lacks (Section~\ref{sec:rotation}). The differences between the columns of Table~\ref{tab:stability} may therefore partly reflect the positions rather than the transformations.

Kimi K2 Turbo had the lowest mirror error (94.94~cp), about half that of the next model, GPT-5.1 (181.08~cp), and only 35 of its 390 pairs with two ordinary scores were inconsistent. It left 109 mirror pairs unanswered, however, so its consistent share (71.0\%) was below those of DeepSeek Chat (76.0\%) and GPT-5.1 (73.2\%) (Figure~\ref{fig:outcomes}).

\subsubsection{Color Swap and Similarity}\label{sec:semantic}

Because the rules treat the two colors symmetrically and the evaluations are reported from White's perspective, a color swap should negate the evaluation (Equation~(\ref{eq:antisym})). Claude Sonnet 4.5 had the lowest color-swap error (198.78~cp), followed by GPT-5.1 (217.49~cp). On the pairs one move apart, GPT-5.1 had the lowest similarity error (297.99~cp): on average, the size of the change in its evaluation differed from the size of the change in Stockfish's by this amount.

\subsubsection{Format Conversion}\label{sec:formatres}

This task compares the evaluations of the same position given as a FEN string and as a natural-language description. Kimi K2 Turbo had the lowest error (36.22~cp), followed by GPT-5.1 (48.11~cp) and Claude Sonnet 4.5 (62.41~cp). Format conversion had the highest consistent shares of the four transformations, from 54.2\% (Grok 4.1 Fast) to 97.4\% (GPT-5.1) (Figure~\ref{fig:outcomes}). Because the description restates the FEN without loss (Section~\ref{sec:format}), the remaining differences cannot come from missing information.

\subsection{Accuracy}\label{sec:accuracy}

Accuracy measures agreement with an external reference, here Stockfish 16.1 at depth 10 (Table~\ref{tab:accuracy}). Claude Sonnet 4.5 (330.41~cp) and GPT-5.1 (362.17~cp) had the lowest accuracy errors, and Grok 4.1 Fast the highest (735.54~cp). For every model, the error was largest on the original positions of the rotation set (the pawnless positions of Section~\ref{sec:rotation}) and smallest on those of the format-conversion set, which come from the first 8 to 20 plies of Lichess games.

\begin{table}[htbp]
\footnotesize\setlength{\tabcolsep}{3pt}
\caption{Accuracy errors: mean absolute difference (cp) between each model's evaluation and that of Stockfish 16.1 at depth 10, for the original positions of each set (FEN input) and for both positions of each similarity pair; lower is better. Positions with an evaluation above 10,000~cp in magnitude are excluded, which leaves from 286 (Gemini 2.5 Flash, Rotation) to 499 (GPT-5.1, Format) positions per cell, and from 319 to 485 pairs for Similarity. Mean is the unweighted mean of the five columns.\label{tab:accuracy}}
\begin{tabularx}{\linewidth}{lCCCCCCC}
\toprule
\textbf{Model} & \textbf{Color Swap} & \textbf{Rotation} & \textbf{Mirror} & \textbf{Similarity} & \textbf{Format} & \textbf{Mean} & \textbf{Rank} \\
\midrule
Claude Sonnet 4.5 & 340.40 & 452.88  & 337.24 & 407.60 & 113.92 & 330.41 & 1 \\
GPT-5.1           & 339.55 & 619.85  & 359.28 & 380.04 & 112.16 & 362.17 & 2 \\
Kimi K2 Turbo     & 361.29 & 685.99  & 396.14 & 397.08 & 129.98 & 394.09 & 3 \\
DeepSeek Chat     & 383.55 & 729.18  & 448.52 & 464.29 & 188.58 & 442.82 & 4 \\
Gemini 2.5 Flash  & 624.47 & 969.84  & 652.78 & 529.87 & 200.31 & 595.45 & 5 \\
Grok 4.1 Fast     & 609.25 & 1340.41 & 558.75 & 636.35 & 532.94 & 735.54 & 6 \\
\bottomrule
\end{tabularx}
\end{table}

\subsection{Accuracy and Stability}\label{sec:tradeoff}

\begin{figure}[htbp]
\centering
\includegraphics[width=12cm]{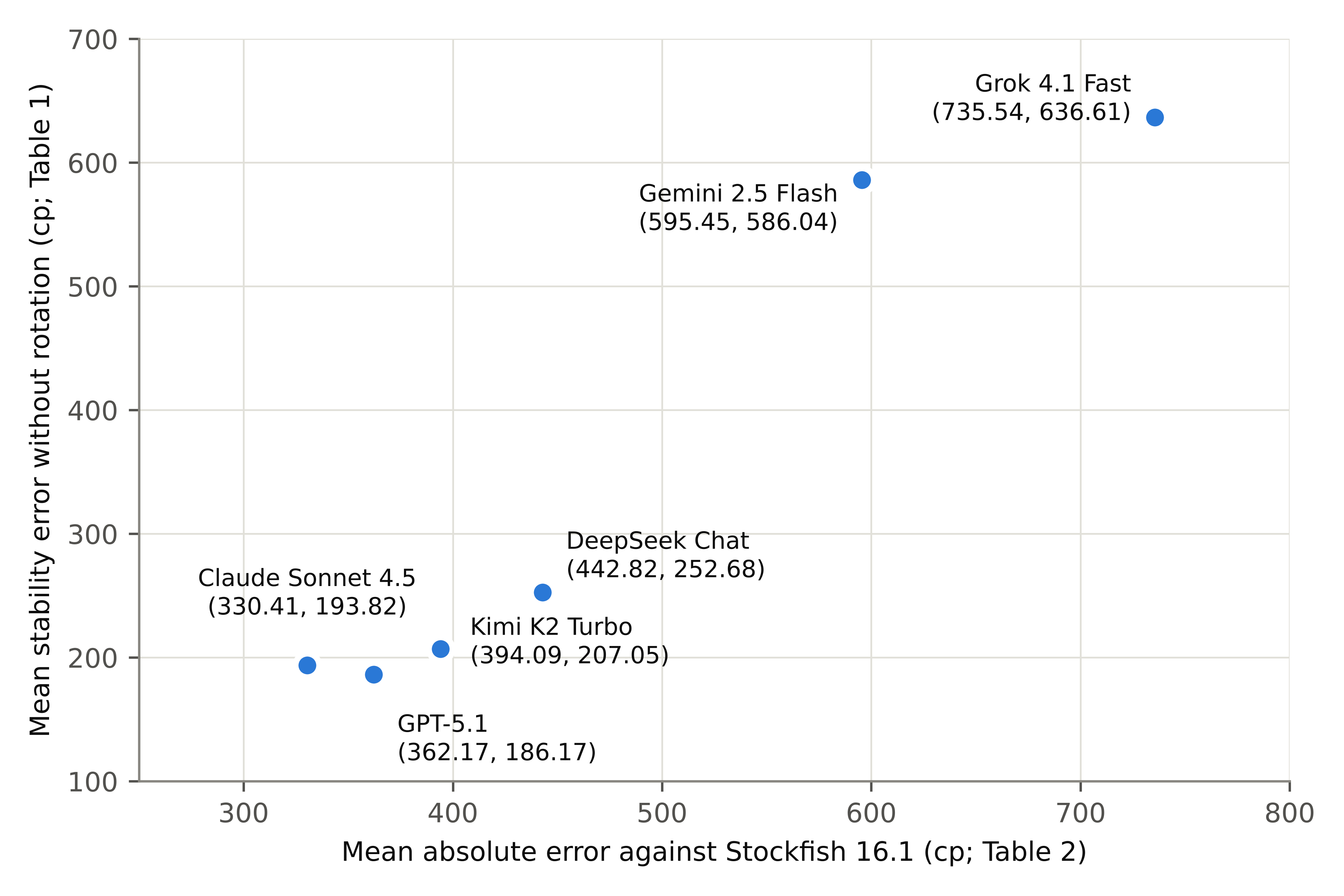}
\caption{Accuracy error against Stockfish 16.1 (Table~\ref{tab:accuracy}) and mean stability error over the color swap, mirror, similarity and format transformations (Table~\ref{tab:stability}, rotation excluded) for the six models; lower is better on both axes. Each label gives the model and its coordinates (accuracy error, stability error) in cp.\label{fig:tradeoff}}
\end{figure}

Figure~\ref{fig:tradeoff} compares the accuracy error with the mean stability error over the four transformations other than rotation. The two measures ranked the models in the same order except for the first two places. GPT-5.1 had the lowest stability error (186.17~cp) and the second-lowest accuracy error (362.17~cp), and Claude Sonnet 4.5 had the second-lowest stability error (193.82~cp) and the lowest accuracy error (330.41~cp). Kimi K2 Turbo and DeepSeek Chat were third and fourth on both measures, and Gemini 2.5 Flash and Grok 4.1 Fast fifth and sixth. With rotation included, GPT-5.1's mean stability error rises to 655.19~cp, the second highest, because its rotation mean contains 352 inconsistent pairs, whereas the other models' rotation means exclude most of their failures (Section~\ref{sec:geom}).

\subsection{Illegal Positions}\label{sec:safety}

An evaluator in an automated system should also reject inputs that are not legal positions. We tested the models on 500 generated illegal positions (e.g., with two White kings or with a White pawn on the first rank). Rejection rates ranged from 44.2\% (DeepSeek Chat) to 96.0\% (Gemini 2.5 Flash) (Table~\ref{tab:safety}). They do not follow the accuracy ranking: Gemini 2.5 Flash, fifth in accuracy, rejected the most illegal positions, whereas DeepSeek Chat returned a numerical evaluation for 279 of the 500. For Kimi K2 Turbo, 41 calls failed because of rate limits; it rejected 74.0\% of all 500 positions and 80.6\% of the 459 it answered. DeepSeek Chat (279) and Grok 4.1 Fast (222) gave numerical evaluations of illegal positions far more often than Gemini 2.5 Flash (20) and GPT-5.1 (50). Our data do not identify the cause; differences in post-training, in reasoning settings (Section~\ref{sec:models}) or in filtering by the provider (Section~\ref{sec:limitations}) could each contribute. Evaluations of impossible positions are a risk in applications that rely on the model to check its input.

\begin{table}[htbp]
\footnotesize\setlength{\tabcolsep}{3pt}
\caption{Responses to the 500 illegal positions per model. A rejection is a \texttt{null} evaluation, as the prompt requires; a numerical evaluation counts as a failure even if the \texttt{error} field names a violation (one answer of Claude Sonnet 4.5). Failed calls returned no parsable answer after four attempts. The rejection rate is the percentage of all 500 positions; higher is better.\label{tab:safety}}
\begin{tabularx}{\textwidth}{lCCCC}
\toprule
\textbf{Model} & \textbf{Rejected} & \textbf{Numerical Evaluation} & \textbf{Failed Call} & \textbf{Rejection Rate (\%)} \\
\midrule
Gemini 2.5 Flash  & 480 & 20  & 0  & \textbf{96.0} \\
GPT-5.1           & 450 & 50  & 0  & 90.0 \\
Claude Sonnet 4.5 & 396 & 104 & 0  & 79.2 \\
Kimi K2 Turbo     & 370 & 89  & 41 & 74.0 \\
Grok 4.1 Fast     & 278 & 222 & 0  & 55.6 \\
DeepSeek Chat     & 221 & 279 & 0  & 44.2 \\
\bottomrule
\end{tabularx}
\end{table}

\subsection{Inconsistent Pairs and Mate-Level Scores}\label{sec:errors}

Section~\ref{sec:stability} measured how much evaluations changed; this section counts the pairs and answers that indicate a failure (Table~\ref{tab:errors}). An inconsistent pair is one whose two ordinary scores differ by more than 300~cp (Section~\ref{sec:metrics}). On the color swap, the mirror and format conversion, from 13.0\% (Claude Sonnet 4.5) to 40.9\% (Grok 4.1 Fast) of the pairs that entered Table~\ref{tab:stability} were inconsistent, so changes of about the value of a minor piece or more occurred for every model even outside rotation.

To check how much these shares depend on the threshold, we varied it from 200 to 500~cp. The order of the four models with the lowest shares (Claude Sonnet 4.5, DeepSeek Chat, Kimi K2 Turbo and GPT-5.1) changed with the threshold, and their shares were within 6.6 percentage points of one another at every threshold tested. Gemini 2.5 Flash and Grok 4.1 Fast had the two highest shares at every threshold. Across the six models, the gaps above 200~cp were not concentrated at large values: 46.5\% of them were at most 400~cp, and 20.2\% exceeded 1000~cp.

\begin{table}[htbp]
\footnotesize\setlength{\tabcolsep}{3pt}
\caption{Error counts by type. Inconsistent pairs: pairs that enter Table~\ref{tab:stability} for the color swap, the mirror and format conversion and whose gap exceeds 300~cp, out of all such pairs. Mate-level scores: numerical answers above 10,000~cp in magnitude, out of all numerical answers, for the color-swap, rotation and mirror sets (both positions of each pair) and for the forced-move set; none of these positions has a mate score at the reference depth. The last two columns refer to the forced-move answers that are not mate-level scores: their accuracy error against Stockfish at depth 12, and the number of them that differ from Stockfish by more than 300~cp.\label{tab:errors}}
\begin{tabularx}{\linewidth}{lCCCCC}
\toprule
 & \textbf{Inconsistent Pairs} & \multicolumn{2}{c}{\textbf{Mate-Level Scores}} & \multicolumn{2}{c}{\textbf{Forced-Move Set, Other Answers}} \\
\cmidrule(lr){3-4}\cmidrule(lr){5-6}
\textbf{Model} & \textbf{w/o Rotation (\%)} & \textbf{Geometric Sets} & \textbf{Forced-Move Set} & \textbf{Accuracy Error (cp)} & \textbf{Error $>$ 300~cp} \\
\midrule
Claude Sonnet 4.5 & 176/1351 (13.0) & 422/2695 & 162/495 & 406.09  & 164/333 \\
DeepSeek Chat     & 204/1436 (14.2) & 334/2728 & 113/494 & 913.03  & 318/381 \\
Kimi K2 Turbo     & 207/1253 (16.5) & 91/2360  & 53/449  & 564.90  & 271/396 \\
GPT-5.1           & 251/1459 (17.2) & 10/2822  & 2/497   & 515.27  & 268/495 \\
Gemini 2.5 Flash  & 390/986 (39.6)  & 578/2376 & 297/437 & 1096.06 & 92/140 \\
Grok 4.1 Fast     & 454/1109 (40.9) & 37/2131  & 29/470  & 1422.86 & 364/441 \\
\bottomrule
\end{tabularx}
\end{table}

Mate-level scores are a second type of failure. The color-swap, rotation, mirror and forced-move sets contain no position with a mate score at the reference depth (Sections~\ref{sec:rotation}--\ref{sec:colorswap} and~\ref{sec:forced}). Gemini 2.5 Flash nevertheless returned a mate-level score for 578 of its 2376 numerical answers on the three geometric sets and for 297 of its 437 numerical answers on the forced-move set. Claude Sonnet 4.5 and DeepSeek Chat did so for 422 and 334 geometric answers and for 162 and 113 forced-move answers, and GPT-5.1 for 10 and 2. Because the prompt asks for values such as $\pm$30,000 for mates (Appendix~\ref{app:prompts}), each of these answers claims a forced mate. Stockfish at depth 12 is not an exact reference. In the authors' earlier re-analysis of the forced-move set at depth 18, with a Stockfish build whose version was not recorded, 224 of the 2179 answers below 10,000~cp concerned a position with a forced mate (Appendix~\ref{app:tactical}). Some mate-level scores on this set may therefore be correct, but none is supported at the reference depth.

On the forced-move set, the accuracy error of the answers that were not mate-level scores ranged from 406.09~cp (Claude Sonnet 4.5) to 1422.86~cp (Grok 4.1 Fast) (Table~\ref{tab:errors}). Appendix~\ref{app:examples} gives one example each of a color-swap failure, of a forced-move position and of a numerical evaluation of an illegal position.

\subsection{Summary of Results}\label{sec:summary}

All six models failed on rotated positions, which they often declared illegal. On the color swap, the mirror and format conversion, 13.0\% to 40.9\% of the pairs with two ordinary scores were inconsistent, and on the four transformations other than rotation the ranking by mean stability error matched the ranking by accuracy except for the first two places, held by GPT-5.1 (stability) and Claude Sonnet 4.5 (accuracy). Gemini 2.5 Flash rejected the most illegal positions (96.0\%) and returned the most mate-level scores for positions without a mate at the reference depth; DeepSeek Chat rejected the fewest illegal positions (44.2\%). Section~\ref{sec:discussion} discusses possible explanations.

\section{Discussion}\label{sec:discussion}

Accuracy against an engine and stability under equivalent transformations measure different properties, but accuracy and the mean stability error over four of the five transformations ranked the six models almost identically (Figure~\ref{fig:tradeoff}). The exception is rotation, on which all six models failed. A $180^\circ$ rotation leaves the value of a pawnless position unchanged but moves the White pieces to the side of the board on which the Black pieces usually stand. Most of the models' \texttt{null} answers declared these legal positions illegal, often by attributing a second king to one side (Section~\ref{sec:geom}). We hypothesize that models trained on large collections of standard game records fit FEN strings in their usual form, which lists the Black pieces' home ranks first and the White pieces' last. The rotated string encodes an equivalent position but is unfamiliar. The models' failure to map it back to the same evaluation, and in many cases to read it correctly, is consistent with a reliance on surface-level token correlations rather than on a model of the board geometry. Outside rotation, inconsistent pairs were still common (13.0\% to 40.9\% of the pairs with two ordinary scores; Table~\ref{tab:errors}). We have no information on whether any provider trained its model on transformed boards.

Mate-level scores for positions without a mate at the reference depth were frequent for Gemini 2.5 Flash, Claude Sonnet 4.5 and DeepSeek Chat and rare for GPT-5.1 (Table~\ref{tab:errors}). Gemini 2.5 Flash returned the most of them, and its API enables thinking by default~\cite{google_thinking}. Of the other five models, GPT-5.1 and Claude Sonnet 4.5 answer without reasoning unless the request asks for it~\cite{openai_gpt51,anthropic_thinking}, the DeepSeek and Grok variants we used are non-reasoning~\cite{deepseek_updates,xai_grok}, and Kimi K2 Turbo is listed separately from the Kimi K2 thinking models~\cite{kimi_quickstart}. Gemini 2.5 Flash was thus the only model that generated reasoning tokens before answering (Section~\ref{sec:models}). On the forced-move set, its answers that were not mate-level scores nevertheless had the second-largest accuracy error (1096.06~cp), larger than that of four of the five models that answered without reasoning. The comparison is confounded by the other differences between the models; a test of the effect of reasoning would need the same model run with and without it.

Numerical evaluations of illegal positions were rarest for Gemini 2.5 Flash (20) and GPT-5.1 (50) and most frequent for DeepSeek Chat (279) and Grok 4.1 Fast (222). Such differences could come from post-training that teaches models to refuse invalid inputs, from reasoning settings or from filters applied by the provider (Section~\ref{sec:limitations}), and our data cannot separate these causes. A \texttt{null} answer is not always a correct rejection: on the rotation set, the models returned 1367 \texttt{null} answers for legal positions, most of which declared the position illegal.

Overall, performance on these logical and spatial tasks did not follow a single ranking: the model that rejected the most illegal positions also returned the most mate-level scores, and the model with the lowest stability error on four transformations failed on rotation like the others. Building symmetries into the model, as geometric deep learning does~\cite{bronstein2021geometric}, adding geometric constraints to training, as surveyed for depth estimation~\cite{vats2025geometric}, or checking outputs against the rules may help to close the gap between pattern matching and grounded evaluation. Long chain-of-thought reasoning~\cite{chen2025longcot} spends additional computation at inference time; Section~\ref{sec:lipschitz} discusses why such computation could matter for positions in which a single move changes the outcome.

\subsection{Limitations}\label{sec:limitations}

All six models were accessed through their providers' APIs. An API can apply undisclosed system prompts, safety filters, post-processing or ensembles that are separate from the underlying transformer. For example, the high rejection rates of Gemini 2.5 Flash and GPT-5.1 for illegal positions could come from an external safety layer or a rule-based pre-filter rather than from the model itself. The differences we observe should therefore be attributed to the behavior of the APIs and not only to the architectures of the models. Without access to the weights and inference stacks we cannot separate a model from the surrounding infrastructure, and this applies to DeepSeek and Kimi as well, because we also queried them through their APIs.

The experimental settings add further limitations. The models did not all run in the same reasoning configuration: Gemini 2.5 Flash reasoned by default, and 144 of its calls reached the limit of 8000 output tokens without an answer, whereas the other five models answered without reasoning (Section~\ref{sec:models}). Each input was evaluated once at temperature 0, so we cannot estimate the variation between repeated calls. Kimi K2 Turbo's 455 failed calls, caused by rate limits and timeouts, reduce its number of answered inputs. Stockfish at depth 10 or 12 is an approximate reference: at such depths it can evaluate equivalent positions differently~\cite{martin2025reevaluating}, and a deeper search finds mates in some forced-move positions (Section~\ref{sec:errors}). The similarity error compares only the size of the change in evaluation, not its direction (Section~\ref{sec:metrics}). The base positions of the similarity set and part of the forced-move positions were reached in games of the Atomic variant (Section~\ref{sec:data}); they are legal in standard chess but may be less typical of standard play than the positions of the format-conversion set.

The rotation set has limitations of its own. Its positions were constructed by removing the pawns from positions reached by random play, so they are legal but were not reached by play. In 223 of the 500 pairs, the original has castling rights that the rotated position lacks, so these pairs are equivalent only within the engine tolerance of 80~cp. Only the $180^\circ$ rotation was tested. The failure of all six models on this set may partly reflect unusual positions rather than the rotation alone, although the same positions before rotation received far fewer \texttt{null} answers (355 against 1367). Rerunning the test on pawnless positions from real games without castling rights, and adding the $90^\circ$ and $270^\circ$ rotations, would separate these factors. Future work could also control for the API factors above with ablation studies on open-weight models (e.g., Llama and Qwen), for which the inference pipeline can be standardized.

\subsection{Consistency Failures as a Breakdown of $G$-Invariance}\label{sec:ginv}

The following formulation describes consistency failures (this subsection) and the evaluation of positions in which a single move changes the outcome (Section~\ref{sec:lipschitz}) in the language of function approximation. It is a framework for interpreting the results, not a model fitted to them.

\textbf{Problem formulation.} Let $\mathcal{S}$ be the set of legal chess positions and $V^*\colon \mathcal{S} \to \mathbb{R}$ the game-theoretic value from White's perspective, with mates mapped to $\pm M$ for a large constant $M$. A board symmetry $g$ maps positions to positions; let $G$ be the group of symmetries that preserve the rules for the positions under test. For positions without castling rights, $G$ contains the left--right reflection $m$ and the color flip $c$ of Section~\ref{sec:colorswap}; $m$ and $c$ commute and generate the Klein four-group $\{e, m, c, mc\}$, in which $mc$ is a $180^\circ$ rotation combined with an exchange of colors. For pawnless positions without castling rights, $G$ also contains the rotations by $90^\circ$, $180^\circ$ and $270^\circ$, the other reflections of the board and the exchange of colors without moving any piece. With evaluations from White's perspective,
\begin{equation}
V^*(g \cdot s) = \chi(g)\, V^*(s) \quad \text{for all } s \in \mathcal{S},\ g \in G,
\label{eq:invariance}
\end{equation}
where $\chi(g) = -1$ if $g$ exchanges the colors and $\chi(g) = +1$ otherwise.

An LLM defines a function $f_\theta\colon \mathcal{T}^* \to \mathbb{R}$ on token sequences, and a position reaches the model through a serialization $\psi\colon \mathcal{S} \to \mathcal{T}^*$, such as FEN. The strings $\psi(s)$ and $\psi(g \cdot s)$ generally differ in their tokens and in the order of those tokens, and nothing in the architecture of $f_\theta$ ties them together.

\textbf{Consistency error.} We define the consistency error as the mean squared violation of Equation~(\ref{eq:invariance}) over the symmetries of a position,
\begin{equation}
\mathcal{E}_{\mathrm{cons}}(\theta) = \mathbb{E}_{s \sim P}\left[\frac{1}{|G|} \sum_{g \in G} \Bigl(f_\theta\bigl(\psi(s)\bigr) - \chi(g)\, f_\theta\bigl(\psi(g \cdot s)\bigr)\Bigr)^2\right],
\label{eq:econs}
\end{equation}
where $P$ is the distribution of test positions. Table~\ref{tab:stability} reports, for each transformation separately, the corresponding mean absolute error over the pairs that enter it, and Figure~\ref{fig:outcomes} the share of pairs whose violation stays within 300~cp.

\textbf{Hypothesis: manifold misalignment.} The rotation results are consistent with models that approximate $V^*$ well on positions as they are usually written (distribution $P$) but not on the induced distribution $P_G = \{g \cdot s \mid s \sim P\}$. We hypothesize that this comes from a reliance on surface-level token patterns rather than on a single latent representation of the position. Let $z_s = \mathrm{Emb}(\psi(s))$ and $z_{g \cdot s} = \mathrm{Emb}(\psi(g \cdot s))$ be the embeddings of the two token sequences. A consistent model would learn a map $h$ with $h(z_s) \approx \chi(g)\, h(z_{g \cdot s})$. Without the inductive biases for group invariance that group-equivariant networks build in~\cite{bronstein2021geometric}, a Transformer may treat the original and the transformed FEN strings as unrelated inputs. The error messages for rotated positions (Section~\ref{sec:geom}), which assign kings to squares and colors they do not occupy, suggest that for this transformation the failure begins with reading the position rather than with evaluating it. We cannot inspect the embedding space, so we cannot test this directly. The results are consistent with the model having learned, in effect, one function $f_{\mathrm{std}}$ for inputs in their usual form and a different, less accurate function $f_{\mathrm{rot}}$ for transformed ones. Analyses of the embeddings or probing classifiers would be needed to distinguish this ``manifold misalignment'' hypothesis, in which the representations of $\psi(s)$ and $\psi(g \cdot s)$ are not mapped to consistent values, from simple overfitting to token $n$-grams.

\subsection{Positions in Which One Move Changes the Outcome: A Lipschitz Perspective}\label{sec:lipschitz}

\textbf{Problem formulation.} In positions where a single move changes the outcome of the game, the game-theoretic value changes sharply between neighboring positions. We analyze such positions with Lipschitz continuity. The argument is general: the forced-move set contains no position with a mate at the reference depth (Section~\ref{sec:forced}), so our experiments do not test it. Let $\mathcal{G} = (\mathcal{S}, \mathcal{E})$ be the undirected graph whose edges join positions that differ by one legal move, with shortest-path distance $d$. For a function $V$ on $\mathcal{S}$, the Lipschitz constant with respect to $d$ equals the largest change across a single edge,
\begin{equation}
\mathrm{Lip}_d(V) = \max_{(u,v) \in \mathcal{E}} |V(u) - V(v)|,
\label{eq:lipgraph}
\end{equation}
because, for any two connected positions $u$ and $v$, the triangle inequality along a shortest path gives
\begin{equation*}
|V(u) - V(v)| \le d(u,v) \max_{(u',v') \in \mathcal{E}} |V(u') - V(v')|,
\end{equation*}
so the ratio $|V(u) - V(v)|/d(u,v)$ never exceeds the maximum over edges, and an edge on which the maximum is attained achieves it. For $V^*$, a single move can change the value by $M$ or more: a move that allows a forced mate turns a drawn position (value 0) into a lost one (value $-M$). Hence $\mathrm{Lip}_d(V^*) \ge M$, and near such positions $V^*$ behaves like a step function.

The network does not see the graph, however; it sees an embedding $x_s \in \mathbb{R}^n$ of the token sequence $\psi(s)$, for example padded token embeddings of a fixed length. If $f$ reproduces $V^*$ at two positions $u$ and $v$, the definition of the Lipschitz constant gives
\begin{equation}
\mathrm{Lip}(f) \ge \frac{|V^*(u) - V^*(v)|}{\|x_u - x_v\|}.
\label{eq:liplower}
\end{equation}
When one move changes only a few characters of the FEN string, $\|x_u - x_v\|$ is small compared with the distances between unrelated positions, whereas $|V^*(u) - V^*(v)|$ can be as large as $M$. Any $f$ that evaluates both positions correctly must therefore have a large Lipschitz constant on these inputs.

\textbf{Spectral bias.} Dot-product self-attention is not Lipschitz on an unbounded input domain~\cite{kim2021lipschitz}, so Transformers have no global Lipschitz bound of the usual form. Networks trained by gradient descent nevertheless show a spectral bias: they fit the low-frequency, smooth components of a target function first~\cite{rahaman2019spectral}. For a feed-forward network $f = W_L \circ \sigma \circ W_{L-1} \circ \cdots \circ \sigma \circ W_1$ with 1-Lipschitz activations $\sigma$, the Lipschitz constant with respect to the Euclidean norm satisfies
\begin{equation}
\mathrm{Lip}(f) \le \prod_{l=1}^{L} \|W_l\|_2,
\label{eq:lipprod}
\end{equation}
where $\|W_l\|_2$ is the spectral norm of the $l$-th weight matrix~\cite{bartlett2017spectrally}. This upper bound, which need not be tight, shows how bounded weight norms limit how sharply such a network can change its output. We hypothesize that, for a model that answers without intermediate reasoning, the computation available per answer limits how well it can approximate the jumps of $V^*$ in regions $\Omega_{\mathrm{tac}}$ where a single move decides the game. The learned function would then be smoother than $V^*$ there and would underestimate decisive advantages.

\textbf{Additional computation.} Suppose that a model's computation consists of $T$ steps, $f_T = h \circ \cdots \circ h$ ($T$ times). Then
\begin{equation}
\mathrm{Lip}(f_T) \le \mathrm{Lip}(h)^T,
\label{eq:lipcomp}
\end{equation}
so the bound on the Lipschitz constant can grow exponentially with $T$. The bound allows, but does not imply, the large Lipschitz constants that Equation~(\ref{eq:liplower}) requires: additional steps of computation remove one obstacle to representing the jumps of $V^*$ without guaranteeing that the model represents them, or that it places them correctly. For Gemini 2.5 Flash such steps need not be latent, because its API generates reasoning tokens by default~\cite{google_thinking} and each generated token requires a further forward pass through the network. In our data, the errors of Gemini 2.5 Flash went in the opposite direction to the smoothing that this argument predicts for models without reasoning: it returned mate-level scores for 297 forced-move positions in which Stockfish found no mate at depth 12 (Section~\ref{sec:errors}). Without access to the model's attention maps or activations, this remains an interpretation of black-box behavior, not a demonstrated property of the architecture.

\section{Conclusions}\label{sec:conclusions}

We evaluated six LLMs on chess positions and on transformed versions of them that a correct evaluator would score identically or with the opposite sign, using metamorphic relations for a $180^\circ$ rotation of pawnless positions, left--right mirroring, color swapping, position pairs one move apart, format conversion and illegal positions. All six models failed on rotation: at most 10.6\% of the 500 rotated pairs received evaluations within 300~cp of each other, and the models often declared the legal rotated positions illegal. On the other four transformations, the ranking by mean stability error matched the ranking by accuracy against Stockfish except for the first two places, held by GPT-5.1 (stability) and Claude Sonnet 4.5 (accuracy). Inconsistent pairs were nevertheless common for every model: 13.0\% to 40.9\% of the pairs with two ordinary scores on the color swap, the mirror and format conversion differed by more than 300~cp. Rejection rates for illegal positions ranged from 44.2\% to 96.0\%. A mean stability error computed only over pairs with two ordinary scores concealed most of the rotation failures, so we recommend reporting the outcome of every query, including refusals and mate-level scores, alongside such means. We refer to the difference between accuracy on positions in their usual form and consistency under transformation as the token--geometry gap. Remedies suggested by our analysis, still to be tested, include training on transformed boards, symmetry-aware architectures or consistency losses, and verification layers that check legality and calculate forced lines before a model answers.

\section*{Acknowledgments}
During the preparation of this manuscript, the authors used Claude (Anthropic; Claude Opus 5.5) to check the analysis code, recompute the results from the raw model outputs, draw the figures, search for related work, check the references, and draft, edit and restructure the text (Section~\ref{sec:genai}). The authors have reviewed and edited the output and take full responsibility for the content of this publication.

\paragraph{Author Contributions.} Conceptualization, Q.H., R.C., W.W. and X.S.; methodology, R.C., W.W. and X.S.; software, R.C., W.W. and X.S.; validation, Q.H. and R.C.; formal analysis, Q.H., R.C., W.W. and X.S.; investigation, R.C. and W.W.; resources, Q.H. and X.S.; data curation, Q.H., R.C., W.W. and X.S.; writing---original draft preparation, Q.H., R.C., W.W. and X.S.; writing---review and editing, Q.H., R.C., W.W. and X.S.; visualization, R.C. and X.S.; project administration, X.S.; funding acquisition, Q.H.

\paragraph{Funding.} Project supported by the Research Foundation of Education Bureau of Hubei Province.

\paragraph{Data Availability Statement.} The original data presented in the study (code, prompts, test sets, model outputs and the scripts that compute the reported results) are openly available in Zenodo at \url{https://doi.org/10.5281/zenodo.24146408}. The Lichess positions come from the Lichess open database~\cite{lichess_db}, which is released under the CC0 license.

\paragraph{Conflicts of Interest.} The authors declare no conflicts of interest. The funders had no role in the design of the study; in the collection, analyses, or interpretation of data; in the writing of the manuscript; or in the decision to publish the results.

\bibliographystyle{unsrtnat_keepcase}
\bibliography{references}

\clearpage
\begin{appendices}
\section{Examples of Errors}\label{app:examples}

\subsection{Color-Swap Failure}\label{app:consistency}

\begin{figure}[htbp]
\centering
\includegraphics[width=13cm]{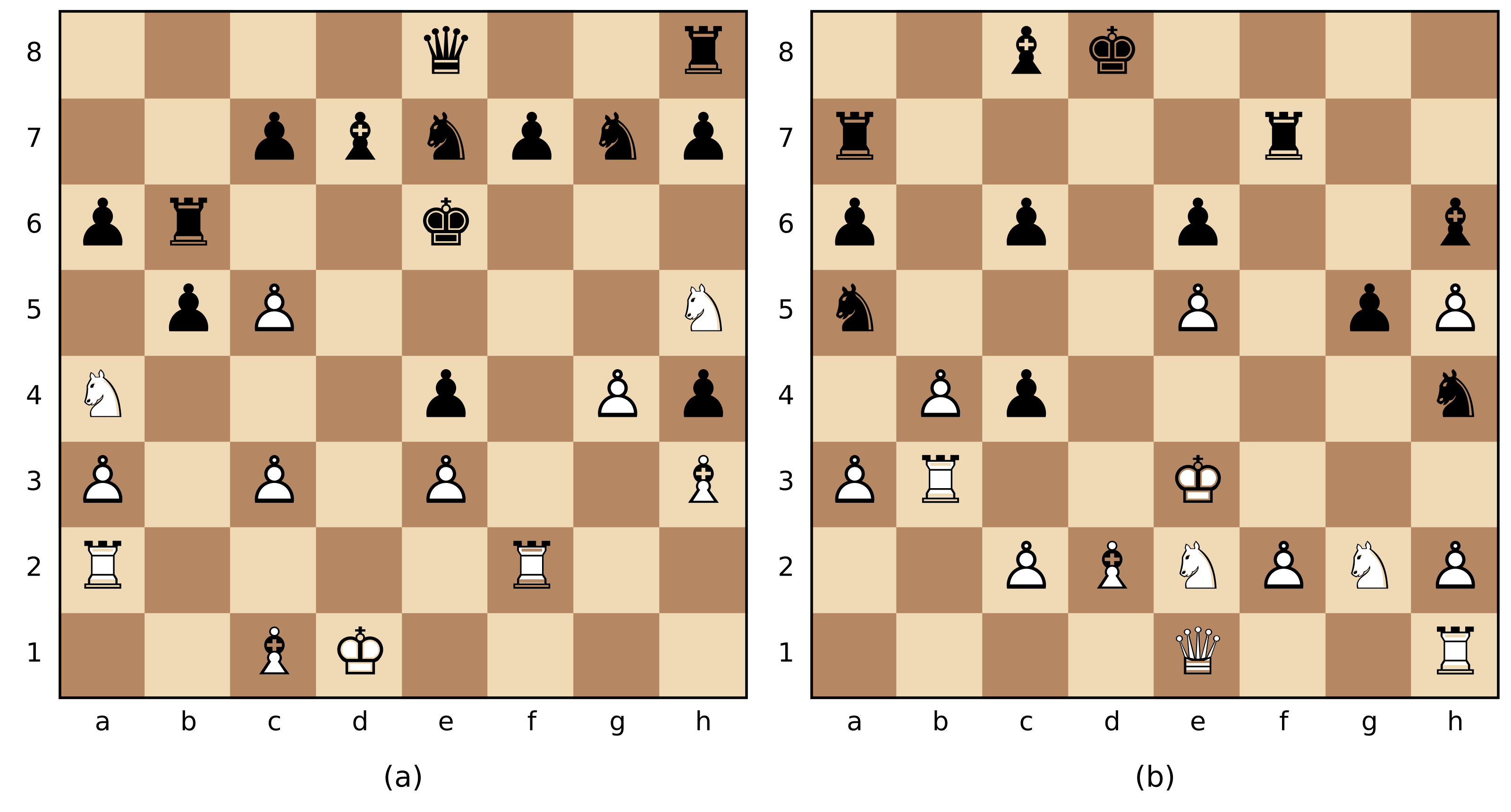}
\caption{Color-swap pair used to illustrate a consistency failure (row 124 of the color-swap set): (\textbf{a}) the original position, Black to move, Stockfish $-447$~cp; (\textbf{b}) the color-swapped position, White to move, Stockfish $+425$~cp. Claude Sonnet 4.5 returned $-450$~cp for both.\label{fig:a1}}
\end{figure}

The position $P$ of Figure~\ref{fig:a1}a, row 124 of the color-swap set,
\begin{center}
\texttt{4q2r/2pbnpnp/pr2k3/1pP4N/N3p1Pp/P1P1P2B/R4R2/2BK4 b - - 1 32},
\end{center}
has Black to move. Its color-swapped counterpart (Figure~\ref{fig:a1}b) is
\begin{center}
\texttt{2bk4/r4r2/p1p1p2b/n3P1pP/1Pp4n/PR2K3/2PBNPNP/4Q2R w - - 1 32}.
\end{center}
Stockfish evaluates $P$ at $-447$~cp and $T_{\mathrm{col}}(P)$ at $+425$~cp from White's perspective, which satisfies Equation~(\ref{eq:antisym}) within the 60~cp tolerance of the set (Section~\ref{sec:colorswap}). Claude Sonnet 4.5 returned $-450$~cp for both positions, with a win probability of 0.15 for White in both. Its evaluation of $P$ agrees with Stockfish's, but for $T_{\mathrm{col}}(P)$, in which White has the advantage that Black had in $P$, it kept the sign, so Equation~(\ref{eq:gaps}) gives $\Delta_{\pm} = 900$~cp. Kimi K2 Turbo did the same ($-280$~cp for both), whereas DeepSeek Chat returned $-150$ and $+150$~cp, which satisfies Equation~(\ref{eq:antisym}) exactly although both magnitudes are smaller than Stockfish's. The response format contains no explanation, so the outputs cannot show why the two models kept the sign.

\subsection{A Forced-Move Position}\label{app:tactical}

\begin{figure}[htbp]
\centering
\includegraphics[width=7cm]{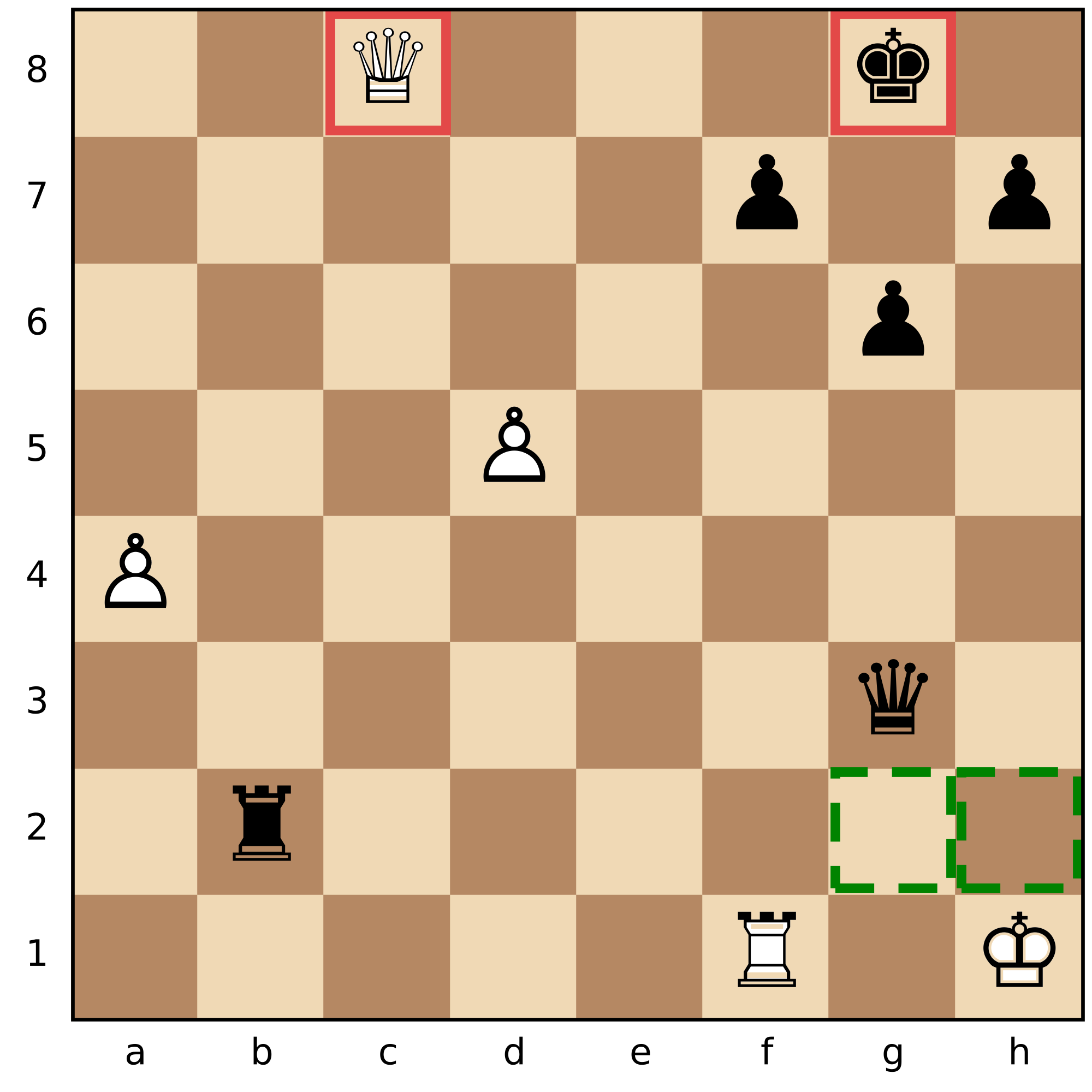}
\caption{Forced-move position (row 187 of the forced-move set). Black is in check from the White queen on c8 (solid red frames on c8 and g8). After the only legal reply, 1...Kg7, Black threatens mate on g2 and h2 (dashed green frames) with the queen on g3 and the rook on b2. Stockfish: $-637$~cp at depth 12 and mate in eight for Black at depth 18; Claude Sonnet 4.5: 0~cp.\label{fig:a2}}
\end{figure}

In the position of Figure~\ref{fig:a2}, row 187 of the forced-move set,
\begin{center}
\texttt{2Q3k1/5p1p/6p1/3P4/P7/6q1/1r6/5R1K b - - 5 36},
\end{center}
Black is in check from the queen on c8 and has one legal reply, 1...Kg7. After it, every check that White can give with the queen (Qf8+, Qg8+, Qh8+ and Qc3+) can be answered by capturing the queen, and Rxf7+ can be answered by capturing the rook. Meanwhile, Black threatens mate on g2 and h2 with the queen on g3 and the rook on b2 (...Qg2\#, ...Qh2\# and ...Rh2\#). Stockfish scored the position at $-637$~cp at depth 12, the depth used to build the set, and found a mate in eight moves for Black at depth 18 in the authors' later re-analysis (Section~\ref{sec:errors}).

The six models returned $-30{,}000$~cp (Gemini 2.5 Flash and DeepSeek Chat), $-1560$~cp (Grok 4.1 Fast), $-900$~cp (GPT-5.1 and Kimi K2 Turbo) and 0~cp (Claude Sonnet 4.5). Claude Sonnet 4.5's score, that of a balanced position, is wrong under either reference, whereas the two mate-level scores, which are unsupported at depth 12, agree with the deeper search. The example shows why the mate-level scores on this set cannot all be counted as errors.

\subsection{Numerical Evaluation of an Illegal Position}\label{app:hallucination}

\begin{figure}[htbp]
\centering
\includegraphics[width=7cm]{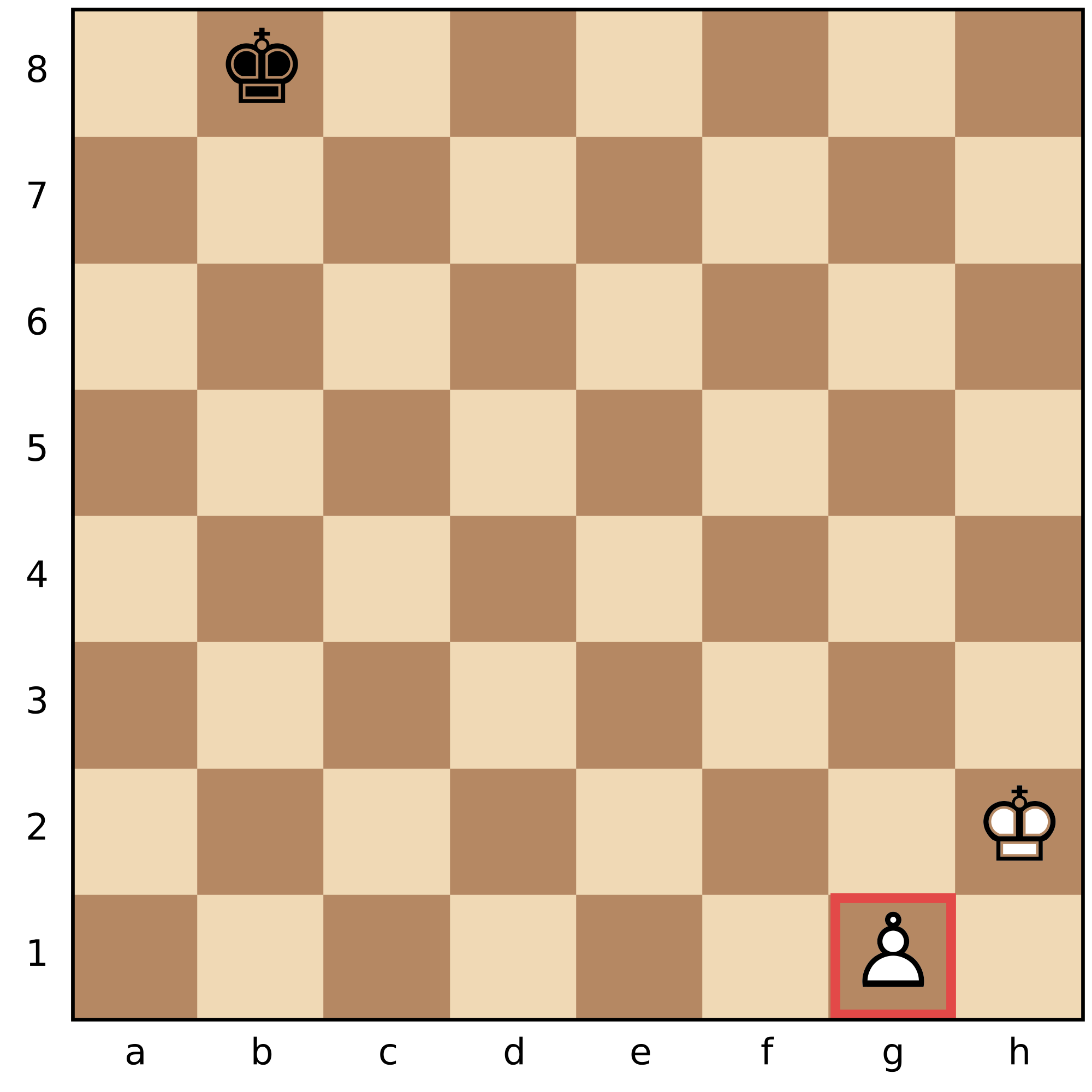}
\caption{Illegal position used to illustrate a numerical evaluation of an illegal position: a White pawn on g1 (solid red frame). DeepSeek Chat: 10,000~cp without an error message; Gemini 2.5 Flash: rejected.\label{fig:a3}}
\end{figure}

The position of Figure~\ref{fig:a3}, row 324 of the illegal set,
\begin{center}
\texttt{1k6/8/8/8/8/8/7K/6P1 w - - 89 156},
\end{center}
has a White pawn on g1. White pawns start on the second rank and move only forward, so a White pawn can never stand on the first rank, and the position cannot arise in a legal game; python-chess reports it as invalid. DeepSeek Chat returned
\begin{center}
\texttt{\{"centipawn\_white": 10000, "win\_probability\_white": 1.0, "error": null\}},
\end{center}
a winning evaluation for White without an error message, and the answer counts as a numerical evaluation of an illegal position (Table~\ref{tab:safety}). GPT-5.1 ($+250$~cp), Claude Sonnet 4.5 (0~cp) and Grok 4.1 Fast ($+950$~cp) also returned numbers. Gemini 2.5 Flash rejected the position with the message ``Illegal position: White pawn on the first rank (g1).'', and the call to Kimi K2 Turbo failed. An engine is no reference for such input: given this FEN, Stockfish returned $+511$~cp in the authors' depth-18 re-analysis, although the position has no game-theoretic value.

\section{Prompts}\label{app:prompts}

All models received the same two templates as the user message, after the system message ``You estimate chess engine-like evaluations and must answer strictly in JSON.''; for Claude Sonnet 4.5, the system message also gave the output schema (Section~\ref{sec:models}). In the templates, \texttt{\{fen\}} is the FEN string, \texttt{\{description\}} the natural-language description and \texttt{\{turn\}} the side to move (\texttt{white} or \texttt{black}). Line breaks were added for display.

\subsection{FEN Input}

Figure~\ref{fig:b1} shows the template used for every input given as a FEN string.

\begin{figure}[htbp]
\begin{Verbatim}[frame=single,fontsize=\footnotesize]
You are a careful chess evaluation assistant. Follow this internal workflow
before you answer:
1. Validate the FEN. Check for illegal conditions: multiple kings of the
  same color, pawns on first/eighth rank, impossible castling rights,
  invalid en passant, or any other rule violations. If the position is
  illegal, you MUST set both `centipawn_white` and `win_probability_white`
  to null and describe the violation in `error`. Do not evaluate illegal
  positions.
2. If the position is legal, identify material balance, king safety, and
  forcing threats from the side-to-move perspective.
3. Convert your evaluation to White's perspective. The output must always be
  from White's point of view, regardless of which side is to move or how the
  board appears (rotated, mirrored, or color-swapped).
After finishing the checklist silently, respond strictly as JSON with this
schema:
{
  "centipawn_white": <number or null>,
  "win_probability_white": <number or null>,
  "error": <string | null>
}
Value specifications:
- `centipawn_white`: Evaluation in centipawns (1 centipawn = 0.01 pawns)
  from White's perspective. Positive values indicate White advantage,
  negative values indicate Black advantage. Typical range: -10000 to +10000.
  For mate positions, use large values (e.g., +30000 for White mate, -30000
  for Black mate).
- `win_probability_white`: White's probability of winning, as a decimal
  between 0.0 and 1.0. 1.0 means White wins, 0.0 means Black wins. This
  should reflect the overall winning chances, not just material balance.
- `error`: Error message string if the position is illegal or evaluation is
  impossible, or null if there is no issue.
Rules:
- Output MUST be a single JSON object with double-quoted keys, no markdown
  or prose.
- The first character of your response must be `{` and the last must be `}`;
  do not add any surrounding text.
- Keep any reasoning internal; do not add extra keys or commentary.
- If the position is illegal, both numeric fields MUST be null and `error`
  must describe the violation.
- If the position is legal but you are uncertain, set numeric fields to null
  and explain briefly in `error` (use null for `error` when there is no
  issue).

FEN: {fen}
Side to move: {turn}
\end{Verbatim}
\caption{Prompt template for FEN input.\label{fig:b1}}
\end{figure}

\subsection{Natural-Language Input}

Figure~\ref{fig:b2} shows the template used for the natural-language inputs of the format-conversion set.

\begin{figure}[htbp]
\begin{Verbatim}[frame=single,fontsize=\footnotesize]
You are a careful chess evaluation assistant. A chess position is described
in natural language below.
Follow this internal workflow before you answer:
1. Validate that the description defines a legal position. Check for rule
  violations: multiple kings of the same color, pawns on first/eighth rank,
  impossible castling rights, or any other illegal conditions. If the
  position is illegal, you MUST set both `centipawn_white` and
  `win_probability_white` to null and describe the violation in `error`. Do
  not evaluate illegal positions.
2. If the position is legal, reconstruct the board mentally, identify
  material balance, king safety, and forcing threats from the side-to-move
  perspective.
3. Convert your evaluation to White's perspective. The output must always be
  from White's point of view, regardless of which side is to move.
After finishing the checklist silently, respond strictly as JSON with this
schema:
{
  "centipawn_white": <number or null>,
  "win_probability_white": <number or null>,
  "error": <string | null>
}
Value specifications:
- `centipawn_white`: Evaluation in centipawns (1 centipawn = 0.01 pawns)
  from White's perspective. Positive values indicate White advantage,
  negative values indicate Black advantage. Typical range: -10000 to +10000.
  For mate positions, use large values (e.g., +30000 for White mate, -30000
  for Black mate).
- `win_probability_white`: White's probability of winning, as a decimal
  between 0.0 and 1.0. 1.0 means White wins, 0.0 means Black wins. This
  should reflect the overall winning chances, not just material balance.
- `error`: Error message string if the position is illegal or evaluation is
  impossible, or null if there is no issue.
Rules:
- Output MUST be a single JSON object with double-quoted keys, no markdown
  or prose.
- The first character of your response must be `{` and the last must be `}`;
  do not add any surrounding text.
- Keep any reasoning internal; do not add extra keys or commentary.
- If the position is illegal, both numeric fields MUST be null and `error`
  must describe the violation.
- If the position is legal but you are uncertain, set numeric fields to null
  and explain briefly in `error` (use null for `error` when there is no
  issue).

Side to move: {turn}
Natural language description:
{description}
\end{Verbatim}
\caption{Prompt template for natural-language input.\label{fig:b2}}
\end{figure}

\subsection{Example of a Natural-Language Description}\label{app:nlexample}

Figure~\ref{fig:b3} shows the FEN of row 0 of the format-conversion set and the description generated from it, which the model received in place of \texttt{\{description\}}.

\begin{figure}[htbp]
\begin{Verbatim}[frame=single,fontsize=\footnotesize]
FEN: r3kbnr/pppb1ppp/2n5/1B2p1B1/4P3/2P2N2/PP3PPP/RN1q1RK1 w kq - 0 9

It's White's turn. White pieces: King g1; Rook f1, a1; Bishop g5, b5; Knight
f3, b1; Pawn e4, c3, h2, g2, f2, b2, a2. Black pieces: King e8; Queen d1;
Rook h8, a8; Bishop f8, d7; Knight g8, c6; Pawn h7, g7, f7, c7, b7, a7, e5.
Castling rights: Black retains kingside castling rights; Black retains
queenside castling rights. Halfmove clock: 0, Fullmove number: 9.
\end{Verbatim}
\caption{FEN and generated natural-language description of the same position.\label{fig:b3}}
\end{figure}

\section{Outcome Counts}\label{app:outcomes}

Table~\ref{tab:outcomes} gives the counts shown in Figure~\ref{fig:outcomes}.

\begin{table}[htbp]
\footnotesize\setlength{\tabcolsep}{3pt}
\caption{Outcome of the 500 pairs of each transformation (Section~\ref{sec:metrics}). Consistent and inconsistent pairs together are the pairs that enter Table~\ref{tab:stability}. Unanswered pairs are split into those with one and with both evaluations \texttt{null} or failed.\label{tab:outcomes}}
\begin{tabularx}{\linewidth}{lCCCCCC}
\toprule
 & & & & \multicolumn{2}{c}{\textbf{Unanswered}} & \\
\cmidrule(lr){5-6}
\textbf{Model} & \textbf{Consistent} & \textbf{Inconsistent} & \textbf{Mate-Level} & \textbf{One} & \textbf{Both} & \textbf{Consistent (\%)} \\
\midrule
\multicolumn{7}{l}{\textit{Color swap}} \\
GPT-5.1 & 355 & 127 & 0 & 14 & 4 & 71.0 \\
Claude Sonnet 4.5 & 377 & 82 & 24 & 17 & 0 & 75.4 \\
Gemini 2.5 Flash & 161 & 137 & 67 & 119 & 16 & 32.2 \\
DeepSeek Chat & 387 & 84 & 10 & 14 & 5 & 77.4 \\
Kimi K2 Turbo & 250 & 172 & 2 & 56 & 20 & 50.0 \\
Grok 4.1 Fast & 170 & 190 & 3 & 116 & 21 & 34.0 \\
\midrule
\multicolumn{7}{l}{\textit{Rotation}} \\
GPT-5.1 & 14 & 352 & 9 & 110 & 15 & 2.8 \\
Claude Sonnet 4.5 & 16 & 32 & 274 & 144 & 34 & 3.2 \\
Gemini 2.5 Flash & 18 & 32 & 178 & 234 & 38 & 3.6 \\
DeepSeek Chat & 8 & 24 & 258 & 195 & 15 & 1.6 \\
Kimi K2 Turbo & 45 & 42 & 68 & 277 & 68 & 9.0 \\
Grok 4.1 Fast & 53 & 51 & 9 & 246 & 141 & 10.6 \\
\midrule
\multicolumn{7}{l}{\textit{Mirror}} \\
GPT-5.1 & 366 & 120 & 0 & 12 & 2 & 73.2 \\
Claude Sonnet 4.5 & 326 & 75 & 24 & 74 & 1 & 65.2 \\
Gemini 2.5 Flash & 136 & 138 & 82 & 125 & 19 & 27.2 \\
DeepSeek Chat & 380 & 89 & 9 & 21 & 1 & 76.0 \\
Kimi K2 Turbo & 355 & 35 & 1 & 87 & 22 & 71.0 \\
Grok 4.1 Fast & 214 & 140 & 1 & 107 & 38 & 42.8 \\
\midrule
\multicolumn{7}{l}{\textit{Format}} \\
GPT-5.1 & 487 & 4 & 0 & 9 & 0 & 97.4 \\
Claude Sonnet 4.5 & 472 & 19 & 5 & 4 & 0 & 94.4 \\
Gemini 2.5 Flash & 299 & 115 & 9 & 71 & 6 & 59.8 \\
DeepSeek Chat & 465 & 31 & 0 & 4 & 0 & 93.0 \\
Kimi K2 Turbo & 441 & 0 & 0 & 42 & 17 & 88.2 \\
Grok 4.1 Fast & 271 & 124 & 0 & 102 & 3 & 54.2 \\
\bottomrule
\end{tabularx}
\end{table}
\end{appendices}

\end{document}